\documentclass[lettersize,journal]{IEEEtran}
\usepackage{amsmath,amsfonts}

\usepackage{algorithmic}
\usepackage{algorithm}
\usepackage{array}
\usepackage{textcomp}
\usepackage{stfloats}
\usepackage{url}
\usepackage{verbatim}
\usepackage[dvips]{graphicx}

\usepackage{cite}
\usepackage{subfigure}
\usepackage{xcolor}
\usepackage{hyperref}

\newcommand{\bvec}[1]{\mbox{\boldmath $#1$}}

\begin{document}

\title{Automatic Extraction of Road Networks by using Teacher-Student Adaptive Structural Deep Belief Network and Its Application to Landslide Disaster}
\author{Shin Kamada and Takumi Ichimura
\thanks{S. Kamada is with Hiroshima City University, Hiroshima, Japan}
\thanks{T. Ichimura is with Prefectural University of Hiroshima, Hiroshima, Japan}
}

\markboth{IEEE Journal of Selected Topics in Applied Earth Observations and Remote Sensing, ~Vol.~16, 2023}%
{Shell \MakeLowercase{\textit{et al.}}: A Sample Article Using IEEEtran.cls for IEEE Journals}


\maketitle

\begin{abstract}
  An adaptive structural learning method of Restricted Boltzmann Machine (RBM) and Deep Belief Network (DBN) has been developed as one of prominent deep learning models. The neuron generation-annihilation algorithm in RBM and layer generation algorithm in DBN make an optimal network structure for given input during the learning. In this paper, our model is applied to an automatic recognition method of road network system, called RoadTracer. RoadTracer can generate a road map on the ground surface from aerial photograph data. A novel method of RoadTracer using the Teacher-Student based ensemble learning model of Adaptive DBN is proposed, since the road maps contain many complicated features so that a model with high representation power to detect should be required. The experimental results showed the detection accuracy of the proposed model was improved from 40.0\% to 89.0\% on average in the seven major cities among the test dataset. In addition, we challenged to apply our method to the detection of available roads when landslide by natural disaster is occurred, in order to rapidly obtain a way of transportation. For fast inference, a small size of the trained model was implemented on a small embedded edge device as lightweight deep learning. We reported the detection results for the satellite image before and after the rainfall disaster in Japan. \textcolor{red}{This version of the article was improved the search algorithm at the border around image.}
\end{abstract}



\section{Introduction}
\label{sec:Introduction}
Recently there have been more cases of extreme climate events including unexpected and unusual weather. The attention of these events has been received in the last few years, due to the significant loss of human lives and escalating economic costs, as well as the impacts on landslides and changes in ecosystems.

In Japan, the Japan Meteorological Agency (JMA) has issued ``Climate Change Monitoring Report'' every year informing the latest status of climate change. According to \cite{Japan}, during the Heavy Rain Event of July 2018, Japan experienced unprecedented heavy rainfall. Overall precipitation observed at AMeDAS stations throughout Japan in July 2018 was extremely high in comparison with past heavy rainfall events since 1982. A prominent characteristic of this rain event is that the record-breaking local precipitation, particularly within 48 to 72 hours, was observed extensively over western Japan and Tokyo region, including the Seto Inland Sea side of Chugoku and Shikoku regions. Due to this heavy rain, river flooding, flood damage, and sediment disasters occurred in many areas, mainly in western Japan, resulting in a catastrophic disaster with more than 250 deaths. In addition, lifelines such as water supply and communications damaged, and traffic obstacles occurred over a wide area.

Due to the disruption of major roads and railroads, the supply was also suspended. Residents shared information on railroads and available roads by posting them on SNS, and searched for moving paths, including routes, while connecting narrow mountain roads. The minimum necessary supplies were transported using these roads, and the temporary roads were constructed instantaneously but it took more than one year for the network to be fully repaired. In local cities, there are few candidates for moving paths between cities, and traffic congestion can be seen in daily life. If the route is disrupted by a disaster, a secondary disaster may occur as a matter of course. Therefore, an automatic detection method of available roads is expected to rapidly obtain a way of transportation and rescue residents when a disaster is occurred.

In recent years, the deep learning \cite{Bengio09} has shown remarkable achievement for various kinds of task as the beginning with the image recognition. The Convolutional Neural Networks (CNN) such as AlexNet \cite{AlexNet}, VGG16 \cite{VGG16}, GoogLeNet \cite{GoogLeNet}, and ResNet \cite{ResNet} showed high image recognition ability. While the CNNs are typical deep learning models, Deep Belief Network (DBN) is also a popular deep learning model, which is the stochastic model by hierarchically stacking the pre-trained Restricted Boltzmann Machines (RBMs) \cite{Hinton06, Hinton12}.
Although these deep learning models have high level of recognition capability, the complicated architectures with large amounts of parameters cause slow inference speed. Therefore, lightweight deep learning models \cite{Wang22} can be also useful in realistic problems using embedded edge devices, where the popular methods to make a small model are pruning, quantization, knowledge distillation, and neural architecture search, and so on.

In the remote sensing fields, the method of automatically recognizing roads from satellite images and aerial photographs is steadily increasing in accuracy by using deep learning techniques, since building a road map and its maintenance require a lot of cost by human experts \cite{Mattyus17, Lian20}. RoadTracer \cite{Bastani18} is an automatic extraction method of road maps using the graph network in addition to the image recognition. The iterative graph search algorithm is conducted to find network graph which represents the connectivity between roads. This is, the algorithm determines whether a vertex and another vertex is connected or not.

However, the image recognition power of RoadTracer is not high since a simple CNN is used for the inference. In addition, long calculation time is required in the graph search because local loop is likely to be occurred in the search. Therefore, the original RoadTracer is improved to enhance the image recognition power and the efficiency of search algorithm in this paper. For the image recognition, the adaptive structural learning method of DBN (Adaptive DBN) \cite{Kamada18_Springer} is used in the inference of deep learning instead of the CNN. The Adaptive DBN has an outstanding function to find the optimal network structure for given input by the neuron generation-annihilation, and layer generation algorithms \cite{Kamada16_SMC, Kamada16_ICONIP, Kamada16_TENCON}. Moreover, a novel method using the Teacher-Student (TS) based Adaptive DBN is proposed. The TS model is an ensemble learning model with one parent and multiple child models to represent ambiguous features. For the graph search algorithm, we often found the cases that the graph search algorithm stopped early because the local loop was occurred in the search algorithm and then a new road was not able to find. To solve the problem, the taboo search is implemented to enhance the ability of global exploration. Finally, our challenge is to propose the detection method of available roads with the satellite images where the natural disaster was actually occurred. As the lightweight deep learning, a small version of our model was developed to remove unnecessary neurons \cite{Kamada20_TENCON, Kamada22_Springer}, and then faster inference with only using CPU was realized on a small embedded PC; Jetson Xavier NX.

The contributions of this paper are as follows. Firstly, the proposed TS model is evaluated on the major image benchmark dataset CIFAR-10/100 \cite{CIFAR10}. The 10-fold cross validation test showed 99.7\% classification accuracy for CIFAR-10 and 95.5\% for CIFAR-100, which were the highest values among the several CNNs and the traditional Adaptive DBN. Secondly, the novel method of RoadTracer using TS based Adaptive DBN is proposed. The detection accuracy of the proposed model was improved from 40.0\% to 89.0\% on average in the seven major cities among the test dataset. Finally, we challenge to apply our proposed model to a local region where the disaster was actually occurred in Japan for the detection of available roads. Our model was applied to two sets of satellite images before and after the disaster. From the difference between two detection results, the available roads were extracted on Jetson Xavier NX.

The remainder of this paper is organized as follows. In the section \ref{sec:adaptive_dbn}, the basic idea of Adaptive DBN is briefly explained. In the section \ref{sec:ts}, the algorithm of TS model and the numeric evaluation using CIFAR-10/100 are explained. In the section \ref{sec:roadtracer}, the basic behavior of searching algorithm in RoadTracer is described. The proposed RoadTracer using TS based Adaptive DBN is proposed and the effectiveness of the model is verified on the satellite images in seven cities. The section \ref{sec:disaster} explains the detection method of available roads in landslide by natural disaster as an application of the proposed model. In the section \ref{sec:discussion}, some discussions were given to compare the effectiveness of our model with the existing remote sensing techniques such as Land use land cover (LULC) classification. This paper is concluded in the section \ref{sec:conclusion}.

\section{Adaptive Structural Learning Method of DBN}
\label{sec:adaptive_dbn}
The Adaptive DBN has been developed as one of the original deep learning model which extents the traditional RBM and DBN. The overview of the method is described as the following sections. 

\subsection{RBM and DBN}
\begin{figure}[tbp]
\centering
\includegraphics[scale=0.5]{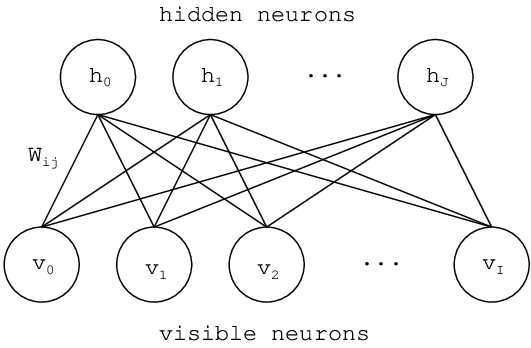}
\caption{Network structure of RBM}
\label{fig:rbm}
\vspace{-5mm}
\end{figure}

An RBM \cite{Hinton12} is a generative model with the visible neurons $\bvec{v} \in \{0, 1 \}^{I}$ for an input and the hidden neurons $\bvec{h} \in \{0, 1 \}^{J}$ to represent feature vectors in data set. The model has the three kinds of learnable parameters $\bvec{\theta}=\{\bvec{b}, \bvec{c}, \bvec{W} \}$, where $\bvec{b} \in \mathbb{R}^{I}$ for the visible neurons, $\bvec{c} \in \mathbb{R}^{J}$ for the hidden neurons, and $W_{ij}$ for the connection between visible and hidden neurons, respectively. As a notable structure of RBMs, only the connections between visible and hidden neurons are available as shown in Fig.~\ref{fig:rbm}. RBM as statistical machine learning gives an energy function $E(\bvec{v}, \bvec{h})$ which represents the distribution for given input as Eq.~(\ref{eq:energy}).

\begin{equation}
E(\bvec{v}, \bvec{h}) = - \sum^{I}_{i} b_i v_i - \sum^{J}_j c_j h_j - \sum^{I}_{i} \sum^{J}_{j} v_i W_{ij} h_j ,
\label{eq:energy}
\end{equation}
\begin{equation}
p(\bvec{v}, \bvec{h})=\frac{1}{Z} \exp(-E(\bvec{v}, \bvec{h})),
\label{eq:prob}
\end{equation}
\begin{equation}
Z = \sum_{\bvec{v}} \sum_{\bvec{h}} \exp(-E(\bvec{v}, \bvec{h})),
\label{eq:z}
\end{equation}
where Eq.~(\ref{eq:prob}) shows the probabilistic model of $\exp(-E(\bvec{v}, \bvec{h}))$. $Z$ shows the energy for all possible patterns of visible and hidden neurons. Typically, the optimal parameters $\bvec{\theta}=\{\bvec{b}, \bvec{c}, \bvec{W} \}$ for given input are estimated by Contrastive Divergence (CD-$k$) \cite{Hinton02}, which is the fast Gibbs sampling method.

Deep Belief Network (DBN) \cite{Hinton06} has a hierarchical network structure. The accumulation of the multiple pre-trained RBMs in order from input to output layer and the fine adjustment of tuning them enables the DBN to represent data distribution with higher accuracy than the RBM. DBN training implements to set the calculated output pattern at the $l-1$-th RBM as an input pattern at the $l$-th RBM.

\subsection{Neuron Generation/Annihilation Algorithm of RBM}
\label{subsec:adaptive_rbm}
The search of both the optimal size of network structure and its parameters is very difficult task for AI designers, because their optimal combination depends on the features to learn the given data. For the problem, the adaptive structural learning methods in RBM and DBN, called Adaptive RBM and Adaptive DBN \cite{Kamada18_Springer} have been developed. The proposed learning method enables the model to generate a new hidden neuron and a new hidden layer according to the training situation. 

\begin{figure}[tbp]
\centering
\includegraphics[scale=0.6]{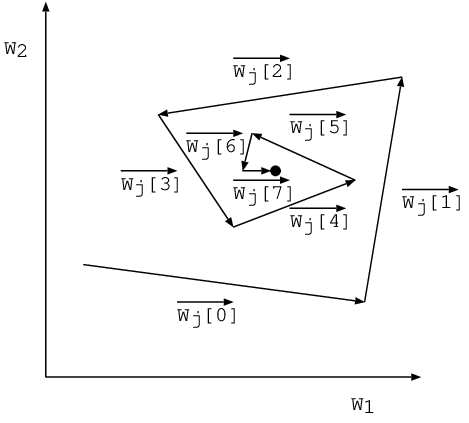}
\vspace{-3mm}
\caption{Convergence situation of Walking Distance (WD)}
\label{fig:WD}
\vspace{-5mm}
\end{figure}

The key idea of the neuron generation in Adaptive RBM is to monitor how the learning parameters is changed in the training epochs, where the amount of change is called Walking Distance (WD) as shown in Fig.~\ref{fig:WD}. Generally, WD will be fluctuated largely after a certain training is progressed if the RBM does not have enough neurons to classify them sufficiently. The situation shows that some hidden neurons may not represent an ambiguous pattern due to the lack of the number of hidden neurons.

\begin{figure}[tbp]
\begin{center}
\subfigure[Neuron generation]{\includegraphics[scale=0.5]{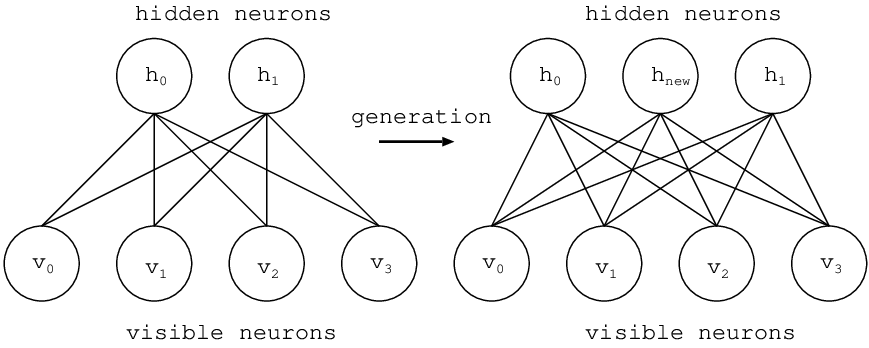}\label{fig:neuron_generation}}
\subfigure[Neuron annihilation]{\includegraphics[scale=0.5]{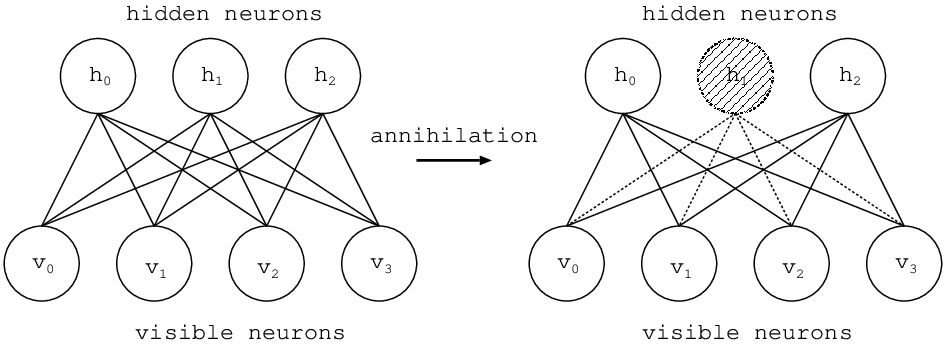}\label{fig:neuron_annihilation}}
\vspace{-3mm}
\caption{Adaptive RBM}
\label{fig:adaptive_rbm}
\vspace{-5mm}
\end{center}
\end{figure}

\begin{figure*}[tbp]
\centering
\includegraphics[scale=0.8]{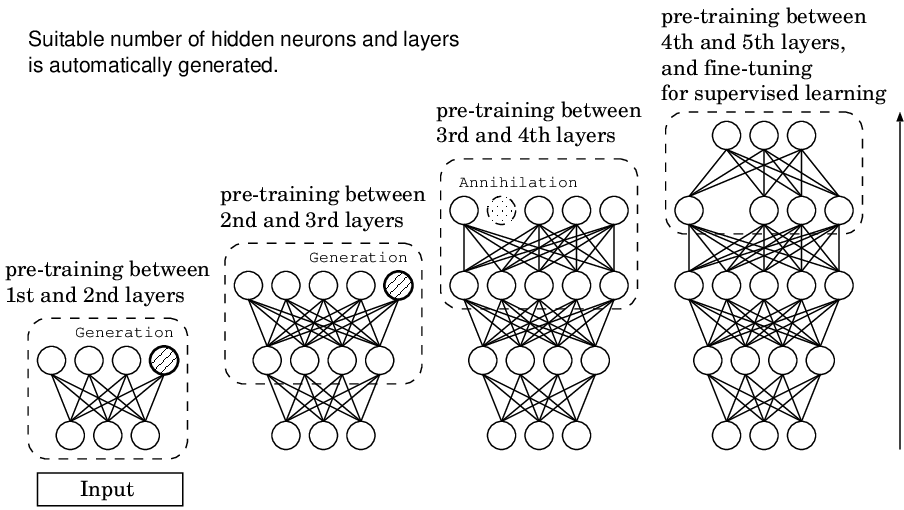}
\vspace{-3mm}
\caption{An overview of Adaptive DBN. The suitable number of hidden neurons is obtained for each pre-training of Adaptive RBM, and the suitable number of hidden layers is also generated by Adaptive DBN.}
\label{fig:adaptive_dbn}
\vspace{-5mm}
\end{figure*}

Therefore, we define the WD as inner product of the amount of changes for $\bvec{c}$ and $\bvec{W}$ except $\bvec{b}$, since noise data in the input space causes the oscillation of $\bvec{b}$. The neuron generation algorithm generates a new neuron if Eq.(\ref{eq:neuron_generation}) is satisfied in the training.
\begin{equation}
\label{eq:neuron_generation}
dc_{j} \cdot dw_{j} > \theta_{G},
\end{equation}
\begin{eqnarray}
\label{eq:WD_c}
dc_{j} &=&  \gamma_{c} dc_{j} + (1 - \gamma_{c}) (|c_j[\tau] - c_j[\tau-1]|),\\
\label{eq:WD_W}
dw_{j} &=&  \gamma_{W} dw_{j} + (1 - \gamma_{W}) ||\bvec{W}_j[\tau] - \bvec{W}_j[\tau-1]||,
\end{eqnarray}
where $dc_{j} (\geq 0)$ and $dw_{j}(\geq 0)$ are the variances for a hidden neuron $j$ during training. $dc_{j} \cdot \theta_{G} ( > 0)$ is a threshold, which is a certain small value. The neuron generation process has been often occurred if $\theta_G$ is a smaller value. The new neuron is inserted to inherit the attributes of the parent hidden neuron as shown in Fig.~\ref{fig:neuron_generation}, to represent ambiguous patterns by two neurons. Please see the original paper \cite{Kamada18_Springer} for the detailed algorithm. As a small revision, Eq.(\ref{eq:neuron_generation}) in \cite{Kamada18_Springer} used two coefficients for $dc_{j}$ and $dw_{j}$ to tune the values, but they are not currently removed because there is no significant difference.

The situation that too much neurons are generated and they become unnecessary or redundant for output due to the parameter setting of $\theta_G$, is assumed. Therefore, the neuron annihilation condition for the hidden neuron activation was proposed as Eq.(\ref{eq:neuron_annihilation}).
\begin{equation}
\label{eq:neuron_annihilation}
\frac{1}{N}\sum_{n=1}^{N} p(h_j = 1 | \bvec{v}_{n}) < \theta_{A},
\end{equation}
where $p(h_j = 1 | \bvec{v}^{(n)})$ means a conditional probability of $h_j \in \{ 0, 1 \}$ for given input data $\bvec{v}^{(n)}$. $\theta_{A}$ is a threshold in $[0, 1]$. The corresponding neuron will be annihilated as shown in Fig.~\ref{fig:neuron_annihilation} if Eq.(\ref{eq:neuron_annihilation}) is satisfied after the neuron generation in the training. The neuron annihilation process has been often occurred if $\theta_A$ is a higher value.

\subsection{Layer Generation Algorithm of DBN}
\label{subsec:adaptive_dbn}
DBN is a hierarchical model of stacking the several pre-trained RBMs. For building hierarchical process, output (activation of hidden neurons) of $l$-th RBM can be seen as the next input of $l+1$-th RBM. Generally, DBN with many RBMs has higher power of data representation than one RBM. Such hierarchical model can represent the various features from an abstract concept to concrete representation at each layer in the direction of input layer to output layer. However, the optimal number of RBMs depends on the target data space.

Adaptive DBN can automatically adjust an optimal network structure to add a new RBM layer one by one based on the idea of WD. If both WD and the energy function do not become small values, a new RBM will be generated to make the network structure suitable for the data set, since the RBM has lacked data representation capability to figure out an input patterns. Therefore, the condition for layer generation is defined by using the total WD at the $l$-th layer, $WD^{l}$, and the energy function, $E^{l}$ as Eq.(\ref{eq:layer_generation1}) and Eq.(\ref{eq:layer_generation2}). 
\begin{equation}
\sum_{l=1}^{k} WD^{l} > \theta_{L1},
\label{eq:layer_generation1}
\end{equation}
\begin{equation}
\sum_{l=1}^{k} E^{l} > \theta_{L2}, 
\label{eq:layer_generation2}
\end{equation}
where $WD^{l} = \sum_{j=1}^{J} (WD_{c_{j}}^{l} \cdot WD_{W_{j}}^{l})$. $WD_{c_{j}}^{l}$ and $WD_{W_{j}}^{l}$ means the value of $WD$ for $c_{j}$ and $\bvec{W}_{j}$ in the $l$-th RBM, respectively. $E^{l}$ is the energy function. $\theta_{L1}$ and $\theta_{L2}$ are the pre-defined threshold values. Fig.~\ref{fig:adaptive_dbn} shows the overview of layer generation in Adaptive DBN (For details see \cite{Kamada18_Springer}). Firstly, the suitable number of hidden neurons is obtained during the pre-training at the first RBM by the neuron generation-annihilation algorithm. Secondly, new Adaptive RBM layers are generated and trained until the layer generation conditions are satisfied for each RBM layer. Finally, the fine-tuning is conducted for a supervised learning task.

\section{Ensemble Learning of Teacher-Student based Adaptive DBN}
\label{sec:ts}
In this section, the proposed ensemble learning of Adaptive DBN is briefly explained. The basic algorithm using Teacher-Student model is described and the numeric experiments using the benchmark datasets CIFAR-10 / 100 \cite{CIFAR10} are conducted to verify the effectiveness of the proposed model.

\subsection{Teacher-Student model}
\label{subsec:ts}
Adaptive DBN has been shown high classification accuracy for large image datasets. However, the model was not able to recognize some confusion and uncertainly cases which include subjective and ambiguous features or inconsistency patterns where input signals are similar but teacher signals are difference. Medical data or human's emotion data are such typical cases, because the annotator's decision for given input will have no small effect on his/her background knowledge. For example, AffectNet \cite{AffectNet} is a facial expression image dataset and eight kinds of emotions are annotated, but its agreement accuracy is not high because human emotion includes subjective information. In the other words, the decrease of classification performance is caused by the uncertainly of annotators' decision process.

An ensemble learning with multiple sub-models is more effective for such confusion cases than a single model. In \cite{Ichimura22_IIAI}, the teacher-student based ensemble learning method of Adaptive DBN (TS model) was proposed for the AffectNet, where the model consists of a parent DBN as teacher and some child DBNs as students. The teacher model tries to train the overall features for the given input data, and each student model tries to train some confusion cases that the parent model was not able to classify them. After the ensemble training, the generated neurons in the student models are transferred to the teacher model as knowledge distillation according to the KL divergence between them. The paper reported the classification accuracy for the AffectNet was improved from 87.4\% to 92.5\% by TS model because the model was able to acquire a new knowledge to distinguish some confusion cases.

\begin{figure*}[tbp]
  \centering
  \includegraphics[scale=3.0]{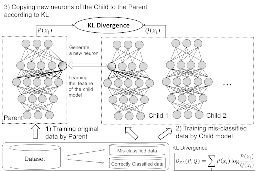}
  \caption{The ensemble learning model of Adaptive DBN. After training the original data by the teacher model, two or more student models are generated and trained for mis-classified samples. Finally, the trained student model's neurons are copied to the teacher model according to the KL divergence as knowledge distillation.}
  \label{fig:KL_abstract}
\end{figure*}

Fig.~\ref{fig:KL_abstract} shows an overview of ensemble learning of TS model. {\bf Algorithm \ref{alg:ts}} shows the algorithm of the model. As shown in Fig.~\ref{fig:KL_abstract}, the algorithm mainly consists of three procedures: 1) a given input data is trained by the Adaptive DBN and it is defined as the teacher model (line 1 in {\bf Algorithm \ref{alg:ts}} and procedure 1) in Fig.~\ref{fig:KL_abstract}), 2) two or more student models are constructed to express mis-classification cases that the teacher model was not able to classify (lines 2-4 in {\bf Algorithm \ref{alg:ts}} and procedure 2) in Fig.~\ref{fig:KL_abstract}), and 3) the student model's neurons are copied to the teacher model according to the KL divergence between them to improve the accuracy (lines 5-6 in {\bf Algorithm \ref{alg:ts}} and procedure 3) in Fig.~\ref{fig:KL_abstract}).

The detailed procedures are explained as follows. Let the teacher model be $T$ as parent, the student model be $S$ as child. Firstly, in the procedure A), the teacher model $T$ trains a given input data by Adaptive DBN using the self-organizing algorithm described in the section \ref{sec:adaptive_dbn}. Secondly, in the procedure B), the multiple student models $S_s$ are initialized by copying the weights from the teacher $T$. The student models $S_s$ train some confusion cases that the parent $T$ was not able to classify. After the student models $S_s$ improved the classification accuracy outstandingly, one student model $S$ with the highest classification accuracy is selected, and the KL divergence between $T$ and $S$ for the given input $x_{i}$ is calculated by Eq.~(\ref{eq:kl_f}) in the procedure C). KL divergence measures the difference between two distributions $T$ and $S$.
\begin{equation}
\label{eq:kl_f}
D_{KL}(T, S) = \sum_{i} P_{T}(x_{i}) \log \frac{P_{T}(x_{i})}{P_{S}(x_{i})},
\end{equation}
where $P_{T}(x_{i})$ and $P_{S}(x_{i})$ are the discrete probability distribution of output for a given input $x_{i} \in X$ to the models $T$ and $S$, respectively. 

\begin{figure}[tbp]
  \centering
  \subfigure[Teacher network]{\includegraphics[scale=0.75]{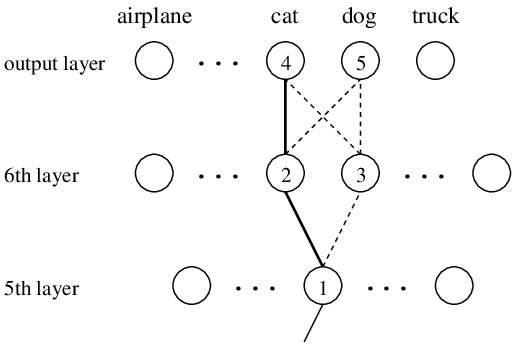}\label{fig:path_parent}}
  \subfigure[Student network]{\includegraphics[scale=0.75]{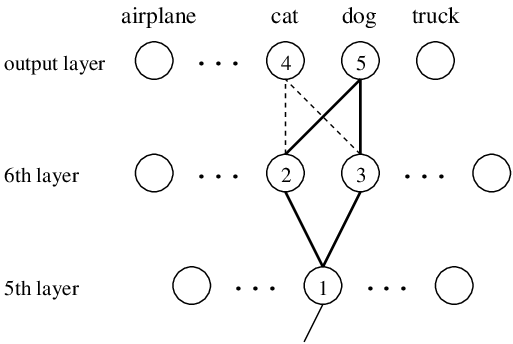}\label{fig:path_child}}
  \caption{Activated neurons and paths}
  \label{fig:result_path_diff}
\vspace{-3mm}
\end{figure}

Finally, some neurons of the student models are transferred to the parent model as the knowledge distillation if the KL divergence $D_{KL}(T, S)$ is larger than the pre-determined threshold $\theta_{KL}$. The large value of $D_{KL}(T, S)$ means that some of neuron's activation patterns for a given input to the two models are different and then it will cause decrease of representation ability for the confusion cases. Fig.~\ref{fig:result_path_diff} shows an example in CIFAR-10 that teacher and student models have different activation neurons and paths for a given input (the detailed explanation will be discussed in the section \ref{subsec:ts_cifar}). Since the student model has higher classification capability for the confusion cases than the teacher model, the improvement of the teacher model can be realized by coping the corresponding neurons of the student model to the teacher model to decrease KL divergence $D_{KL}(T, S)$. After the insertion process, the weights for the generated neurons in the teacher model are fine-tuned with small oscillation to improve the classification accuracy. In this paper, we set $\theta_{KL} = 0.0015$ from the preliminary experiment results \cite{Ichimura22_IIAI}. 

\begin{algorithm}[tbp]
\caption{Algorithm for the ensemble model}
\label{alg:ts}                        
\begin{algorithmic}[1]
  \STATE The Adaptive DBN trains a given data and the model is defined as the teacher model $T$.
  \STATE Sub data sets $SubSet = \{Set_{1}, \cdots, Set_{i}, \cdots, Set_{L} \}$ for the mis-classified cases that the teacher model $T$ was not able to classify, are constructed, where $L$ is the number of sub data sets. In this paper, the sub data set has a certain number of mis-classified cases between categories based on the confusion matrix of training cases.
  \STATE For each sub data set $Set_{i}$, one student model $S_{i}$ is constructed. The student model is initialized by copying the network structure of teacher model $T$.
  \STATE Each student model $S_{i}$ trains the corresponding sub dataset $Set_{i}$.
  \STATE One student model $S$ with the highest classification accuracy in the student models is determined and KL divergence between $T$ and $S$ by Eq.(\ref{eq:kl_f}) is calculated. If the KL divergence is larger than a certain threshold $\theta_{KL}$ for the given data, the neuron's activation path is investigated for the two models. If a different activation path is found, the corresponding neurons of the student model $S$ are copied to the teacher model $T$ to decrease KL divergence $D_{KL}(T, S)$. After the insertion process, the weights for the generated neurons in the teacher model are fine-tuned with small oscillation to improve the classification accuracy.  
  \STATE The algorithm ends when no new neurons are copied to the teacher model. For inference after the training, the student models are no longer used.
\end{algorithmic}
\end{algorithm}

\subsection{Numeric Evaluation on CIFAR-10 / CIFAR-100}
\label{subsec:ts_cifar}
The proposed TS model was applied to CIFAR-10/100 as major image benchmark dataset \cite{CIFAR10}, although the effectiveness of the previous Adaptive DBN was investigated in the paper \cite{Kamada18_Springer}. The dataset has 50,000 and 10,000 cases for training and test, respectively. Each image has $32 \times 32 \times 3$ pixels and is classified to 10 classes for CIFAR-10 and 100 classes for CIFAR-100, respectively. The parameter setting of the Adaptive DBN was as follows: the training algorithm was Stochastic Gradient Descent (SGD), the batch size was 64, the learning rate was 0.005, the initial number of hidden neurons was 400, $\theta_{G} = 0.001$, $\theta_{A}=0.100$, $\theta_{L1}= 0.05$, and $\theta_{L2}= 0.05$, as reported in the paper \cite{Kamada18_Springer}. For the TS model, $\theta_{KL} = 0.0015$ and the number of student models for the TS model was compared with 5 and 10.

Table~\ref{tab:result_cifar10} shows the classification accuracy on CIFAR-10. For the comparison, the results of some CNN models \cite{Goodfellow13, Clevert16, Benjamin15, Zagoruyko16} and the Adaptive DBN with/without TS were shown. Moreover, 10-fold cross validation test was conducted for only the Adaptive DBN. Although the original Adaptive DBN \cite{Kamada18_Springer} had higher classification accuracy than the CNNs, some confusion cases for five classes `bird', `cat', `dog', `automobile', and `truck' were found in the Adaptive DBN. Therefore, five student models were constructed for the classes by the proposed TS model. As the result, the TS model improved 2.3\% classification accuracy compared to the original Adaptive DBN.

Fig.~\ref{fig:result_path_diff} shows an example of the activated neurons and paths for given input to the teacher and student models. A node and a line between nodes are neuron and weight, respectively. The bold line and the dotted line indicate the actually activated path and no activated one. In the case, the teacher model predicted the wrong output as `cat' through 1, 2, and 4 neurons, while the student model predicted the correct output as `dog' because the path from 1 to 3 was activated. In the knowledge distillation, such neurons in the student model are copied to the teacher model and fine-tuned, then the teacher model becomes to predict the correct output as `dog'.

Table~\ref{tab:result_cifar100} shows the classification accuracy on CIFAR-100. As same as the result of CIFAR-10, the Adaptive DBN showed higher classification accuracy than the CNNs. Since CIFAR-100 contains more complex cases to classify than CIFAR-10, 10 student models were constructed in the TS model, where the confusion cases for 10 classes `otter', `seal', `squirrel', `rabbit', `shrew', `lizard', `bear', `woman', `boy', and `willow'. As a result, the TS model achieved the highest classification accuracy among the other models by 14.3\% improvement of accuracy.

\begin{table}[tbp]
\caption{Classification Accuracy on CIFAR-10}
\vspace{-3mm}
\label{tab:result_cifar10}
\centering
\begin{tabular}{l|r|r}
\hline 
\multicolumn{1}{c|}{Model} & \multicolumn{1}{c|}{Accuracy} & Std. \\ \hline
Convolutional NN (MaxOut) \cite{Goodfellow13}            & 0.883 & -  \\ 
Convolutional NN (ELU-Network) \cite{Clevert16}          & 0.934 & - \\ 
Convolutional NN (Fract. Max Pooling) \cite{Benjamin15}  & 0.965 & - \\ 
Convolutional NN (Wide ResNet) \cite{Zagoruyko16}        & 0.960 & - \\ 
Adaptive DBN \cite{Kamada18_Springer}             & {\bf 0.974}& 0.008 \\ 
Adaptive DBN (TS, No. of student models: 5)         & {\bf 0.997}& 0.001 \\
\hline
\end{tabular}
\end{table}

\begin{table}[tbp]
\caption{Classification Accuracy on CIFAR-100}
\vspace{-3mm}
\label{tab:result_cifar100}
\centering
\begin{tabular}{l|r|r}
\hline 
\multicolumn{1}{c|}{Model} & \multicolumn{1}{c|}{Accuracy} & Std.\\ \hline
Convolutional NN (Maxout) \cite{Goodfellow13}            & 0.614 & -  \\ 
Convolutional NN (ELU-Network) \cite{Clevert16}          & 0.757 & -  \\ 
Convolutional NN (Fract. Max Pooling) \cite{Benjamin15}  & 0.723 & -  \\ 
Convolutional NN (Wide ResNet) \cite{Zagoruyko16}        & 0.807 & -  \\ 
Adaptive DBN \cite{Kamada18_Springer}                               & {\bf 0.812} & 0.008 \\ 
Adaptive DBN (TS, No. of student models: 5)                            & {\bf 0.858} & 0.019  \\
Adaptive DBN (TS, No. of student models: 10)                            & {\bf 0.955} & 0.008  \\ 
\hline
\end{tabular}
\end{table}

\section{Automatic Extraction of Road Networks using TS Adaptive Model}
\label{sec:roadtracer}
In the section \ref{subsec:roadtracer_related_works}, the existing works related to the extraction method of road networks such as RoadTracer are explained. The section \ref{subsec:roadtracer_alg} briefly explains the basic algorithm of RoadTracer. In the section \ref{subsec:roadtracer_dbn}, the original RoadTracer is improved to enhance the image recognition power by Adaptive TS model and efficiency of graph search algorithm by the taboo list. To verify the effectiveness of the proposed model, the experiments results for road map detection are conducted in the section \ref{subsec:roadtracer_exe}.

\subsection{Related works}
\label{subsec:roadtracer_related_works}
Recently, the attention of remote sensing using deep learning approaches is highly increased because of high accuracy of deep learning techniques \cite{Zhu2017}. The several applications using high resolution satellite images or related sensing data were developed for the tasks of scene classification, object detection, segmentation, LULC classification, image fusion, image registration, and so on \cite{Lei2019}. For object detection, the paper \cite{Wegner2016} proposed an object detection model of urban trees using both aerial images and street view images, since some deep learning models used two or more sensing data such as hyperspectral, SAR, and LiDAR data, for the improvement of the performance, not just a single image. Wu et.al. proposed a U-Net model which enhances to grasp global and local contrast information for the detection of small objects in infrared images \cite{Wu23}.

For LULC classification, Houston2013 Data Set \cite{houston2013} is a typical benchmark dataset to classify high resolution satellite images with 144 spectral channels to 15 kinds of objects such as tree, soil, road, and so on. For the dataset, Hong et.al. proposed a fusion model using a small size of graph CNN and a general CNN, where the performance of the test data archived 89.47\% \cite{Hong21A}. Recent success model is multi-modal deep learning which recognizes multiple channels of high resolution images as different modalities, where the model archived 92.91\% \cite{Hong21B}. SOT-Net \cite{Zhang1} achieved 93.21\% using hyperspectral image and LiDAR data to effectively extract structural information transmission and physical properties alignment.

Domain adaptation techniques are also developed as cross-scene hyperspectral image classification, where two different domain datasets are provided as source data and target data. The typical task of domain adaptation is classification of target data for given source data. For the problem, TSTnet \cite{Zhang2} has a graph structure of CNN to represent topological structure and semantic information in the cross-scene hyperspectral image. Gia-CFSL \cite{Zhang3} proposed a few-shot learning algorithm for the problem that the number of classes in source data and target data is different. SDEnet \cite{Zhang4} uses the idea of generative adversarial learning (GAN) for domain adaptation. Moreover, the fusion method with not only hyperspectral image but also natural language model was proposed as LDGnet \cite{Zhang5}.

In the remote sensing, the method of automatically recognizing roads from satellite images or aerial photographs is also steadily increasing, since building a road map and its maintenance require a lot of cost by human experts. DeepRoadMapper \cite{Mattyus17} is a segmentation technique using deep learning to automatically detect roads from a satellite image. Deep Window \cite{Lian20} extracts a center point of roads from a small patch image and connects the detected points in the iterative algorithm.

However, these methods were not able to achieve high detection accuracy, since some non-road features such as trees, river, and shadow of buildings covered roads and they became noise \cite{He20}. In the other words, the model has insufficient segmentation power to clearly distinguish the road from the non-road features from only an aerial image. In order to solve the problem, RoadTracer \cite{Bastani18} applied the idea of graph network in addition to the image recognition. The iterative graph search algorithm is conducted to find network graph which represents the connectivity between roads. This is, the algorithm determines whether a vertex and next vertex is connected or not by given satellite image and detected roads around the current searching position.

The other recent methods tackled the modification approaches of connectivity between the detected roads. VecRoad \cite{Tan20} proposed the point-based iterative exploration framework which modifies the connectivity and aliment of the detected roads by the exploration guidance. RoadNet \cite{Liu18} proposed the fusion model of automatic segmentation by the three CNNs and user interaction, where the desired predictions are obtained by user interaction due to the difficulty to learn prior knowledge for the complex road maps. Bahl et.al. proposed a road graph extraction model to train node features, intersection, and these offsets using ResNet \cite{Bahl22}. Zhou et.al, proposed D-LinkNet \cite{Zhou18} using dilated convolution layers for reducing computational cost. Mei et.al. proposed CoANet \cite{Mei21} using attention modules to train connectivity of roads. Zhang et.al. proposed NodeConnect \cite{Zhang23} that simultaneously predicts node confidence and connectivity maps in encoder-decoder architecture, where the model achieved state of the art performance for the RoadTracer dataset (77.96\%).

In this paper, our challenge is to detect available roads from satellite images after a natural disaster is occurred. The methods using pre-defined knowledge are not appropriate for the task because the road map will be dynamically changed due to the disaster then suitable knowledge cannot be pre-defined. Although it may be possible to use only the RoadTracer for the task, the image recognition power is not high because a simple CNN is used for the inference. In addition, long calculation time is required in the graph search because local loop is likely to be occurred in the search. Therefore, the original RoadTracer is improved to enhance the image recognition power by Adaptive TS model and the efficiency of search algorithm by the taboo list in the section \ref{subsec:roadtracer_dbn}.

\subsection{RoadTracer}
\label{subsec:roadtracer_alg}
RoadTracer \cite{Bastani18} is the automatic recognition method of a road map on the ground surface from aerial photograph data. The iterative graph search algorithm is used to find network graph connectivities between roads using graph structure, which is a different approach of segmentation based method of deep learning.

{\bf Algorithm \ref{alg:Ite_Graph_Construction}} shows a pseudo code of the search algorithm. The search algorithm firstly initializes any starting location of search $\bvec{v}_0$, an empty graph $\bvec{G}$ with the size of given satellite image, and a stack of vertices $\bvec{S}$. $\bvec{v}_0$ is a vertex which indicates the coordinates of the given high resolution satellite image. $\bvec{G}$ is used to store detected vertices and edges, where an edge can be seen as road if the corresponding two vertices are connected. The stack $\bvec{S}$ is used to store previous positions in the search algorithm and initialized to push $\bvec{v}_0$. $\bvec{S}_{top}$ is defined by a vertex of top position in the stack $\bvec{S}$ and it shows the current position in the search. 

\begin{figure*}[tbp]
  \centering
  \includegraphics[scale=0.8]{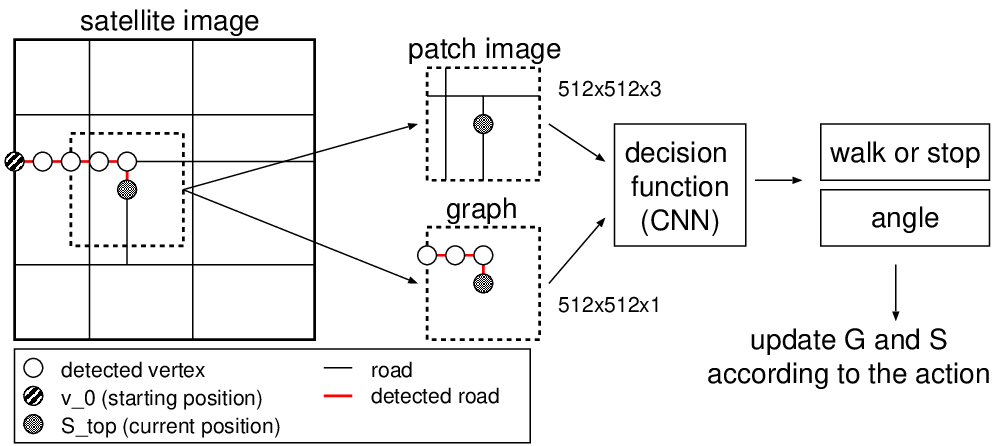}
  \caption{Overview of decision function}
  \label{fig:decisionfunction}
\end{figure*}

Secondly, the decision function takes an action that is walk or stop for the given $\bvec{G}$ and the satellite image around the current position $\bvec{S}_{top}$ and then updates $\bvec{G}$ and $\bvec{S}$ according to the decided action. Fig.\ref{fig:decisionfunction} shows the overview of input and output structure of the decision function. A simple CNN which consists of 17 convolution layers without pooling and full connected layers for the prediction is used as the decision function in the original RoadTracer \cite{Bastani18}. The input of the CNN takes $512 \times 512 \times 4$ data which consists of $512 \times 512 \times 3$ small patch of the satellite image and $512 \times 512 \times 1$ mask of graph $\bvec{G}$ around the current position $S_{top}$ as shown in Fig.\ref{fig:decisionfunction}. For the input, the CNN predicts two kinds of possibilities as output, which are action and angle $\alpha$ (line 3 in {\bf Algorithm \ref{alg:Ite_Graph_Construction}}). The action is defined as `walk' or `stop' by two neurons $O_{action} = (O_{walk}, O_{stop})$, which is calculated by softmax. The angle $\alpha$ is represented by $a$ neurons $O_{angle} = (o_{1},\cdots,o_{i},\cdots,o_{a})$ evenly arranged in $[0, 2\pi]$, which is calculated by sigmoid function. In this paper, $a = 64$. The candidate vertex $\bvec{u}$ is obtained as next position by moving a certain distance $D$ from $S_{top}$ for maximum direction $\rm{arg max}\it{_{i}(o_{i})}$ (line 4 in {\bf Algorithm \ref{alg:Ite_Graph_Construction}}). If $O_{walk}$ is larger than the pre-determined threshold $T$, the algorithm takes `walk' to the candidate vertex $\bvec{u}$, then the algorithm updates $\bvec{G}$ and $\bvec{S}$ (lines 8-10 in {\bf Algorithm \ref{alg:Ite_Graph_Construction}}). Otherwise, `stop' is determined, then the current searching position $\bvec{S}_{top}$ is popped from the stack of vertices $\bvec{S}$ to back the previous position in next step (lines 5-6 in {\bf Algorithm \ref{alg:Ite_Graph_Construction}}). Finally, the searching step is repetitively conducted until the terminate condition is satisfied.

\begin{figure*}[tbp]
  \centering
  \includegraphics[scale=0.7]{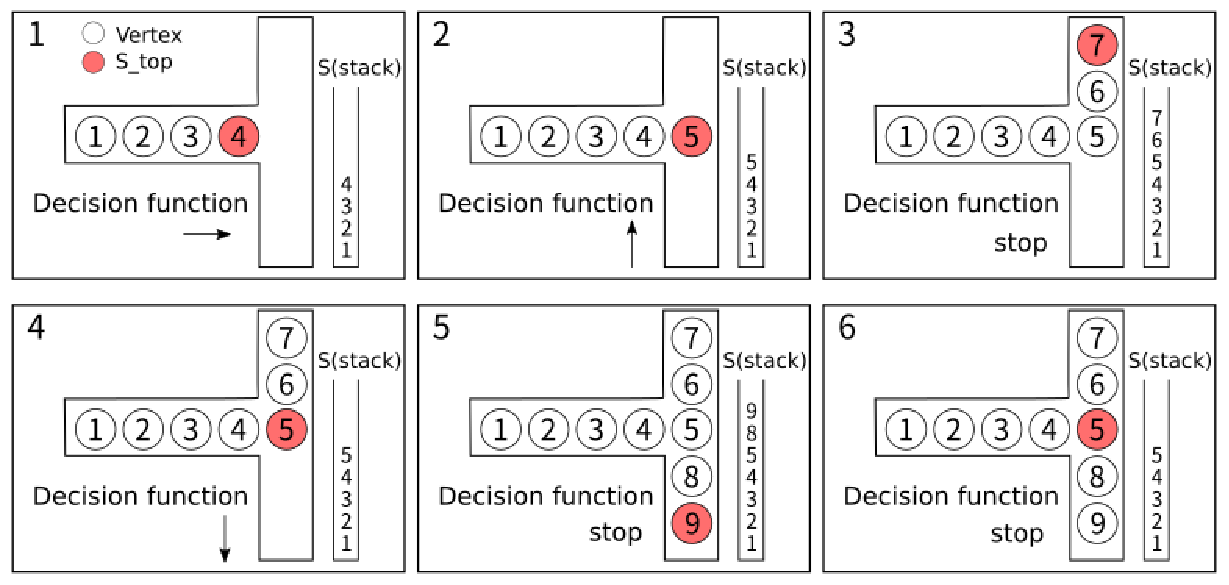}
  \caption{Example of Search Algorithm}
  \label{fig:SearchAlgorithm}
\end{figure*}

Fig.~\ref{fig:SearchAlgorithm} is an example of searching behavior in an intersection. In the case, it is supposed that the vertices 1 to 4 are stacked in $\bvec{S}$. Firstly, the decision function takes `walk' to east direction, then the vertex 5 is pushed to $\bvec{S}$. Secondly, the decision function takes `walk' to north direction as same procedure. Thirdly, the function takes `stop' because the available vertices are not found. Consequently, the current position is moved to back the intersection and the algorithm searches the south direction. Finally, the stop action is determined again, then the search algorithm is finished.

\begin{algorithm}[tbp]
\caption{Graph Search \cite{Bastani18}}
\label{alg:Ite_Graph_Construction}
\begin{algorithmic}[1]
\REQUIRE A satellite Image and a constant value $D$ are given.
\STATE initialize a vertex $\bvec{v}_{0}$ as starting location, an empty graph $\bvec{G}$, and a stack of vertexes $\bvec{S}$ with $\bvec{v}_{0}$. $\bvec{S}_{top}$ is defined by the top position of $\bvec{S}$.
\WHILE {$\bvec{S}$ is not empty}
 \STATE \textit{action}, $\alpha$ := decision func($\bvec{G}$, $S_{top}$, \textit{Image})
 \STATE $\bvec{u}$ := $\bvec{S}_{top} + (D \cos\alpha, D\sin\alpha)$
  \IF {action = stop or $\bvec{u}$ is located outside of \textit{Image}}
  \STATE  pop $\bvec{S}_{top}$ from $\bvec{S}$
  \ELSE
  \STATE  add vertex $\bvec{u}$ to $\bvec{G}$
  \STATE  add edge $(\bvec{S}_{top}, \bvec{u})$ to $\bvec{G}$
  \STATE  push vertex $\bvec{u}$ onto $\bvec{S}$
  \ENDIF
\ENDWHILE
\end{algorithmic} 
\end{algorithm}

\subsection{RoadTracer using Adaptive DBN}
\label{subsec:roadtracer_dbn}
The RoadTracer was implemented on our Adaptive DBN instead of the CNN in order to improve the detection accuracy. Of course, the intermediate hidden layers are automatically constructed for input data by the self-organization function of our Adaptive DBN. The input and output layers of Adaptive DBN are trained based on the specification of RoadTracer as mentioned in the section \ref{subsec:roadtracer_alg}.

In the previous research, the developed RoadTracer using Adaptive DBN had higher detection accuracy than the CNN for the satellite images in both the major 45 cities and the suburban area \cite{Kamada21_ICIEV}. However, some mis-detection results were found in seven of 45 cities, because ambiguous features were contained in the input data. For example, some non-road features such as trees, river, and shadow of buildings covered roads and they became noise. Moreover, the detection accuracy was decreased for the features only included in some cities. In addition, for such cases, the graph search algorithm stopped early because the local loop was occurred in the search and then a new road was not able to find. In this paper, the proposed TS model, as mentioned in the section \ref{sec:ts}, is firstly applied to the RoadTracer for the improvement of image recognition capability. 

Secondly, the taboo search is implemented to the graph search algorithm for the efficiency of the search. In the original RoadTracer, the cases that the search was finished early without searching the entire area in the satellite image, was found. The reason could be that the local loop was occurred around the roads where the model was not able to detect. In the case, if the model cannot detect a road continuously, the search algorithm will be finished. Therefore, {\bf Algorithm \ref{alg:Ite_Graph_Construction}} is slightly changed for implementation of the taboo search. In the previous search algorithm, the next candidate position $\bvec{u}$ is determined by the decision function and walk to $\bvec{u}$ if the $O_{walk}$ is larger than the pre-determined threshold. In the proposed algorithm, if the search algorithm doesn't find a new vertex for $n$ consecutive times, the corresponding vertex is appended to the taboo list as forbidden solution, then the next search will be conducted in the area excluding taboo list. Then the search algorithm can continue move to the end without stopping in the middle.

\subsection{Experiments results}
\label{subsec:roadtracer_exe}
In this section, the proposed taboo search and TS model are evaluated on the satellite images of major cities. In the previous research, the Adaptive DBN was evaluated on 40 cities of satellite images such as Chicago, which is used in the original paper \cite{Bastani18}. In the all 40 cities, the Adaptive DBN achieved higher detection accuracy than the original RoadTracer. However, the detection accuracy of the Adaptive DBN was worse for 7 of 40 cities, which are London, Louisville, New York, Tokyo, Amsterdam, Vancouver, and Montreal, because the search algorithm was finished early not to reach the end by the cut off points in the roads. Therefore, the seven cities are evaluated by the proposed taboo search and TS model in this paper. The satellite images and graph data were obtained from Google Map API and OpenStreetMap \cite{OpenStreetMap}, respectively. The area of each city was about 24 square km and the resolution of the satellite image was 60cm per 1 pixel. The size of input image to the model was $512 \times 512 \times 3$ RGB image. The following GPU workstation was used to the experiments: Nvidia GeForce RTX 3090 $\times 2$, Intel(R) Core(TM) i9-10900K CPU @ 3.70GHz, 64GB RAM. The parameter setting of the proposed model was followed as described in the section \ref{subsec:ts_cifar}. The 10 student models were created according to $\theta_{KL} = 0.0015$. For the RoadTracer, the parameters $a = 64$ and $T = 0.5$ were used.

\begin{table*}[tbp]
\caption{Detection accuracy for seven cities}
\vspace{-3mm}
\label{tab:result_roadtracer_ts}
\begin{center}
\begin{tabular}{l|l|r|r|r|r||r|r|r|r}
\hline
 & & \multicolumn{4}{|c||}{Searching time} & \multicolumn{4}{|c}{Detection Accuracy} \\ \cline{3-10}
City & Model & Ave. & Std. & Max & Min & Ave. & Std. & Max & Min \\ \hline
London & Adaptive DBN & 8310.10 & 1193.68 & 9384 & 5168 & 16.7\% & 0.0264 & 18.8\% & 9.5\% \\
& Adaptive DBN + TB & 31663.40 & 107.05 & 31830 & 31530 & 73.6\% & 0.0345 & 77.9\% & 65.8\% \\
& Adaptive DBN + TB + TS & 37498.30 & 876.74 & 39514 & 36773 & {\bf 85.3\%} & 0.0242 & {\bf 89.8\%} & {\bf 82.7\%} \\\hline
Louisville & Adaptive DBN & 21979.30 & 6413.03 & 31656 & 17680 & 44.7\% & 0.1314 & 64.2\% & 35.5\% \\
& Adaptive DBN + TB & 21663.60 & 593.93 & 22474 & 20278 & 61.0\% & 0.0231 & 65.5\% & 58.7\% \\
& Adaptive DBN + TB + TS & 54742.20 & 890.94 & 56200 & 53642 & {\bf 92.1\%} & 0.0435 & {\bf 98.8\%} & {\bf 84.1\%} \\ \hline
New York & Adaptive DBN & 5808.50 & 1795.10 & 6693 & 907 & 14.1\% & 0.0436 & 17.4\% & 2.4\% \\
& Adaptive DBN + TB & 32067.10 & 578.05 & 33075 & 31093 & 83.0\% & 0.0212 & 85.9\% & 79.5\% \\
& Adaptive DBN + TB + TS & 49845.80 & 784.74 & 51802 & 49065 & {\bf 90.8\%} & 0.0559 & {\bf 98.6\%} & {\bf 82.9\%} \\ \hline
Tokyo & Adaptive DBN & 25103.50 & 15810.10 & 35231 & 1378 & 29.3\% & 0.1913 & 42.0\% & 1.1\% \\
& Adaptive DBN + TB & 53187.50 & 999.13 & 54545 & 50994 & 71.0\% & 0.0442 & 78.6\% & 64.3\% \\
& Adaptive DBN + TB + TS & 64100.30 & 1052.39 & 65182 & 62361 & {\bf 86.4\%} & 0.0286 & {\bf 91.7\%} & {\bf 82.4\%} \\ \hline
Amsterdam & Adaptive DBN & 28505.10 & 1249.82 & 30061 & 26818 & 49.1\% & 0.0280 & 52.1\% & 45.2\% \\
& Adaptive DBN + TB & 38983.00 & 882.59 & 40238 & 37883 & 72.8\% & 0.0384 & 79.9\% & 68.0\% \\
& Adaptive DBN + TB + TS & 50739.90 & 995.05 & 52061 & 48911 & {\bf 88.0\%} & 0.0228 & {\bf 91.5\%} & {\bf 84.5\%} \\ \hline
Vancouver & Adaptive DBN & 28346.70 & 10047.51 & 32993 & 31 & 61.2\% & 0.2175 & 72.1\% & 0.1\% \\
& Adaptive DBN + TB & 36957.00 & 919.96 & 38577 & 35487 & 80.5\% & 0.0421 & 87.8\% & 76.3\% \\
& Adaptive DBN + TB + TS & 47259.80 & 937.39 & 48350 & 45330 & {\bf 88.9\%} & 0.0322 & {\bf 93.0\%} & {\bf 83.1\%} \\ \hline
Montreal & Adaptive DBN & 36081.80 & 16403.24 & 46595 & 718 & 64.6\% & 0.2759 & 81.5\% & 0.6\% \\
& Adaptive DBN + TB & 46031.30 & 1250.75 & 48138 & 44007 & 83.1\% & 0.0639 & 92.9\% & 75.3\% \\
& Adaptive DBN + TB + TS & 55341.60 & 600.30 & 56070 & 54063 & {\bf 91.4\%} & 0.0356 & {\bf 95.9\%} & {\bf 86.2\%} \\ 
\hline
\end{tabular}
\end{center}
\end{table*}

\begin{figure*}[tbp]
  \centering
  \subfigure[London, Adaptive DBN]{\includegraphics[scale=1.20]{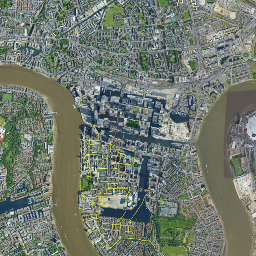}\label{fig:london_adbn}}
  \subfigure[London, Adaptive DBN + TB]{\includegraphics[scale=1.20]{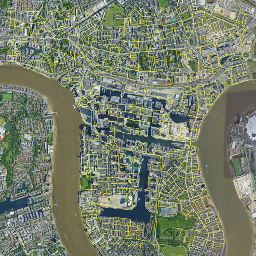}\label{fig:london_tb}}
  \subfigure[London, Adaptive DBN + TB + TS]{\includegraphics[scale=1.20]{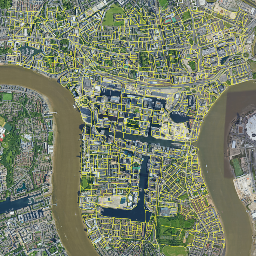}\label{fig:london_ts_tb}}
  \subfigure[Louisville, Adaptive DBN]{\includegraphics[scale=1.20]{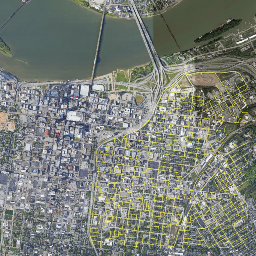}\label{fig:louisville_adbn}}
  \subfigure[Louisville, Adaptive DBN + TB]{\includegraphics[scale=1.20]{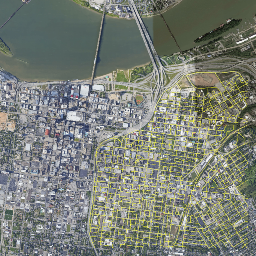}\label{fig:louisville_tb}}
  \subfigure[Louisville, Adaptive DBN + TB + TS]{\includegraphics[scale=1.20]{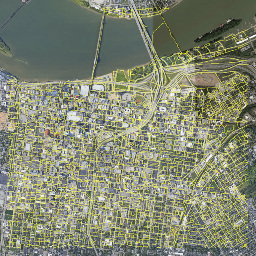}\label{fig:louisville_ts_tb}}
  \caption{Detection results}
  \label{fig:result_roadtracer_ts}
\end{figure*}

The proposed TS model was applied to the above seven cities to improve the detection accuracy. Table \ref{tab:result_roadtracer_ts} shows the searching time and the detection accuracy for the seven cities. The searching time means the number of steps or epochs that the graph search algorithm has been finished. The detection accuracy means the ratio that the ground-truth vertices are actually detected by the model. For the comparison, three methods of Adaptive DBN with/without the taboo search (TB) and the TS model (TS) were evaluated as shown in Table \ref{tab:result_roadtracer_ts}. We tested the experiment 10 trials and the average result was shown to verify the detection performance.

From Table \ref{tab:result_roadtracer_ts}, the detection accuracy of the proposed model was much higher than that of the Adaptive DBN. Since the searching time of Adaptive DBN was much lower with higher variance than that of the proposed model, we consider the Adaptive DBN was not able to detect roads which contain complicated features and then the searching process was finished early. By using the TB search, the searching time was increased with smaller variance and the explorable area was expanded because the search was progressed for the direction excluding forbidden solutions in the taboo list. As a result, the case that the search stopped early was decreased.

Although the TB search improved the detection performance in many cities, the detection accuracy for Louisville was worse because of lack of image recognition performance. Fig.~\ref{fig:result_roadtracer_ts} shows the detection results on London and Louisville for three models. Compared with/without TB search, the Adaptive DBN with TB showed better performance than that without TB for London as shown in Fig.~\ref{fig:london_adbn} and Fig.~\ref{fig:london_tb}. On the other hands, for Louisville as shown in Fig.~\ref{fig:louisville_adbn} and Fig.~\ref{fig:louisville_tb}, no significant improvement in detection accuracy for two models was observed. The maximum detection accuracy was 64.2\% and 65.5\% for the Adaptive DBN without and with TB. The both two models was not able to detect the roads including junctions between highways and general roads, then the left area in Fig.~\ref{fig:louisville_adbn} and Fig.~\ref{fig:louisville_tb} was not searched. Therefore, the TS model trained the complicated cases by the multiple student models and the trained knowledge was transferred into the parent model. As a result, the maximum detection accuracy was improved from 65.5\% to 98.8\%. As shown in Fig.~\ref{fig:louisville_ts_tb}, the TS model was able to detect many roads over the wide roads and bridges in Louisville because such features were trained in the student models. In the all cities, the Adaptive DBN with TB and TS showed the highest detection accuracy among the other models. The reason for the lack of accuracy for London could be that there were no bridges crossing the area in the satellite image.

Table \ref{tab:result_roadtracer_comparision} shows the detection accuracy of the RoadTracer dataset for comparison with the recent models as mentioned in the section \ref{subsec:roadtracer_related_works}. Our Adaptive DBN with TB and TS showed the highest detection accuracy among the other recent models.

\begin{table}[tbp]
\caption{Comparison result with recent models}
\label{tab:result_roadtracer_comparision}
\centering
\begin{tabular}{l|r}
\hline 
\multicolumn{1}{c|}{Model} & \multicolumn{1}{c}{Detection Accuracy} \\ \hline
RoadTracer \cite{Bastani18} & 64.81 \%  \\
D-LinkNet \cite{Zhou18} & 72.16 \%  \\
CoANet \cite{Mei21} & 74.14 \%  \\
NodeConnect \cite{Zhang23} & 77.96 \%  \\
Our model & {\bf 86.29} \%  \\
\hline
\end{tabular}
\end{table}

\begin{figure}[tbp]
\centering
\includegraphics[scale=0.6]{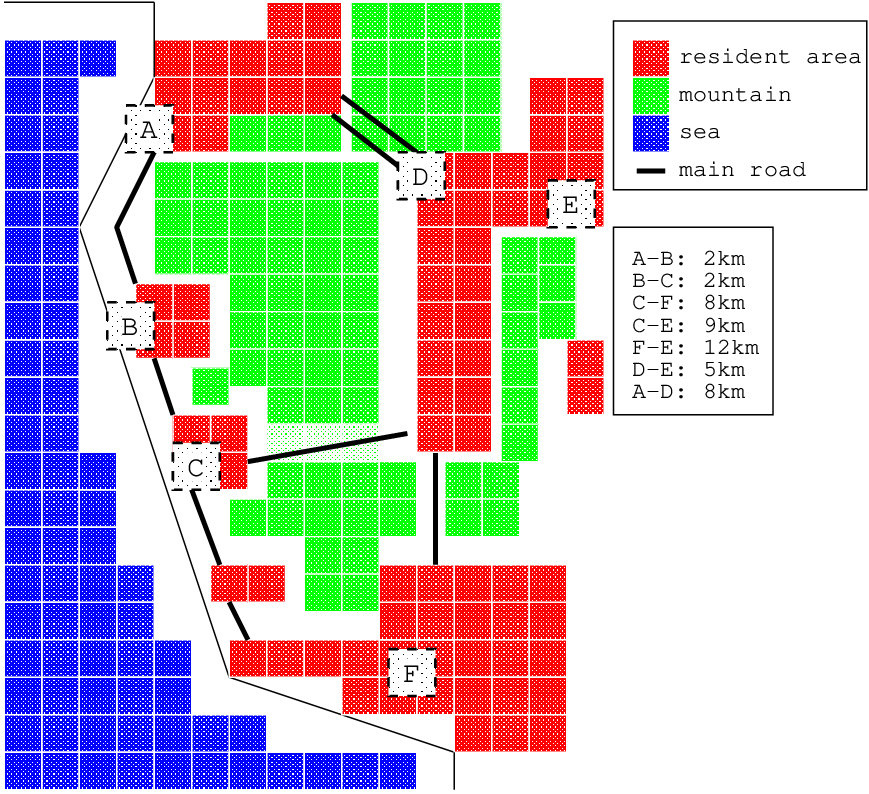}
\vspace{-3mm}
\caption{Rough map around areas A-F}
\label{fig:map}
\vspace{-5mm}
\end{figure}

\begin{figure}[tbp]
\centering
\includegraphics[scale=1.5]{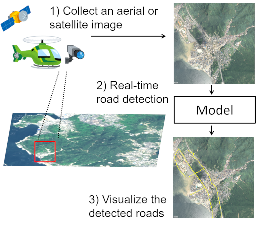}
\vspace{-5mm}
\caption{An example of system implementation}
\label{fig:system_overview}
\vspace{-5mm}
\end{figure}

{\color{red}
\subsection{Enclave search}

As shown in Fig.~\ref{fig:result_roadtracer_ts}, we noticed that, despite the presence of unexplored roads, the areas corresponding to the outer edges of the satellite image had not been explored. This phenomenon was also observed in other cities within the dataset not mentioned in the paper. Although there are still some roads in the edges of the image during the search, the system determined that the road was a dead end and that there were no more explorable roads, so it terminated the search for that path. This issue is caused by a flaw in the search algorithm in the areas at the edges of the image. When attempting to search all the way to the edge of the image, the algorithm tries to search outside the image, resulting in an error and preventing the search from continuing. If there are no more candidates in the search area and no path to the next search area can be found, the system will terminate the search. While this behavior is consistent with the algorithm’s design, if the search encounters a dead end or a river, it has no choice but to terminate because there are no further candidates. As shown in Case 3 and Case 9 in Fig.~\ref{fig:SearchAlgorithm}, while it is possible to perform a search when a branch is required, the search was not performed if there was no region beyond the branch. In other words, if such a region were searched, the regions shown in \textcircled{\scriptsize 7} and \textcircled{\scriptsize 9} would extend beyond the boundaries of the image; however, since they do not exist, the search must be stopped. Although this kind of processing is considered necessary when encountering a dead end, no processing was performed for cases where the image reached its edge. In fact, there were cases where parts of other areas in the image remained unexplored, causing the system to terminate the search midway. This is no different from stopping the search altogether. Even if the system were able to explore points \textcircled{\scriptsize 7} and \textcircled{\scriptsize 9} in Fig.~\ref{fig:SearchAlgorithm}, it would still amount to stopping there. For this reason, when encountering the edge of a region, it is necessary to explore not only the surrounding area but also other, more distant regions. In such cases, a portion of the land is considered an enclave -- a geographically isolated area surrounded by other regions-- and a search algorithm is needed to handle this enclave. Furthermore, performing a search of the surrounding area may lead to regions outside the image or to areas without roads, such as rivers or mountains.

To address this issue, {\bf Algorithm \ref{alg:enclave}} is a method designed to enable continuous road network exploration by incorporating the following processes:
\begin{itemize}
\item ``returning the search point to the position prior to the jump,'' and
\item ``returning to the original location while taking into account the exploration rate of the area to which the jump was made.''
\end{itemize}

\begin{algorithm}[tbp]
\caption{Algorithm for the enclave search}
\label{alg:enclave}                        
\begin{algorithmic}[1]
{\color{red}  
\REQUIRE Let $\bvec{G}$ be the current graph and $\bvec{u}$ be the set of vertices that have been searched. Let $r$ be the threshold.
\STATE Randomly change the search point.
\STATE Perform the search according to {\bf Algorithm \ref{alg:ts}}.
\STATE Let $\bvec{u}'$ be the set of vertices obtained from the search. Let $\lvert \bvec{u}’ \rvert / \lvert \bvec{u} \rvert$ be the proportion of newly detected vertices.
\IF {$\lvert \bvec{u}’ \rvert / \lvert \bvec{u} \rvert >  r$ then }
    \STATE Add the obtained vertex $\bvec{u}'$  and its edges to graph $\bvec{G}$
\ENDIF
\STATE Return to the search point prior to the change.
}
\end{algorithmic}
\end{algorithm}

Following this algorithm, the experimental results obtained using the improved search system are presented in Table \ref{tab:result_roadtracer_ts_enclave} and Fig.~\ref{fig:result_roadtracer_ts_enclave}.

\begin{table*}[tbp]
\caption{Detection accuracy for seven cities}
\vspace{-3mm}
\label{tab:result_roadtracer_ts_enclave}
\begin{center}
\begin{tabular}{l|l|r|r|r|r||r|r|r|r}
\hline
 & & \multicolumn{4}{|c}{Prior Method} & \multicolumn{4}{|c}{Enclave Search} \\ \cline{3-10}
City & Model & Ave. & Std. & Max & Min & Ave. & Std. & Max & Min \\ \hline
London & Adaptive DBN & 16.7\% & 0.0264 & 18.8\% & 9.5\% & 29.1\% & 0.0746 & 35.9\% & 13.8\%\\
& Adaptive DBN + TB & 73.6\% & 0.0345 & 77.9\% & 65.8\% &79.0\% & 0.0136 & 81.3\% & 76.6\%\\
& Adaptive DBN + TB + TS & 85.3\% & 0.0242 & 89.8\% & 82.7\% & {\bf 95.7\%} & 0.0042 & {\bf 96.2\%} & {\bf 94.9\%} \\\hline
Louisville & Adaptive DBN & 44.7\% & 0.1314 & 64.2\% & 35.5\% & 42.1\% & 0.2092 & 51.0\% & 21.2\%\\
& Adaptive DBN + TB & 61.0\% & 0.0345 & 65.5\% & 58.7\% & 77.7\% & 0.0087 & 79.3\% & 76.4\%\\
& Adaptive DBN + TB + TS & 92.1\% & 0.0242 & 98.8\% & 84.1\% & {\bf 93.3\%} & 0.0062 & {\bf 94.1\%} & {\bf 92.0\%}\\ \hline
New York & Adaptive DBN & 14.1\% & 0.0436 & 17.4\% & 2.4\% & 22.1\% & 0.1808 & 54.4\% & 16.9\%  \\
& Adaptive DBN + TB & 83.0\% & 0.0212 & 85.9\% & 79.5\% & 83.1\% & 0.0013 & 84.4\% & 80.9\%  \\
& Adaptive DBN + TB + TS & 90.8\% & 0.0559 & 98.6\% & 82.9\% & {\bf 91.0\%} & 0.0058  & {\bf 91.8\%} & {\bf 90.0\%} \\ \hline
Tokyo & Adaptive DBN & 29.3\% & 0.1913 & 42.0\% & 1.1\% & 27.8\% & 0.1116 & 45.4\% & 20.9\%\\
& Adaptive DBN + TB & 71.0\% & 0.0442 & 78.6\% & 64.3\% & 68.1\% & 0.0254 & 72.3\% & 65.3\%\\
& Adaptive DBN + TB + TS & 86.4\% & 0.0286 & 91.7\% & 82.4\% & {\bf 92.4\%} & 0.0069 & {\bf 93.8\%} & {\bf 91.6\%}\\ \hline
Amsterdam & Adaptive DBN & 49.1\% & 0.0280 & 52.1\% & 45.2\% & 51.2\% & 0.0182 & 68.6\% & 46.1\%\\
& Adaptive DBN + TB & 72.8\% & 0.0384 & 79.9\% & 68.0\% & 82.1\% & 0.0042 & 88.0\% & 79.5\%\\
& Adaptive DBN + TB + TS & 88.0\% & 0.0228 & 91.5\% & 84.5\% & {\bf 92.8\%} & 0.0080 & {\bf 94.0\%} & {\bf 91.5\%}\\ \hline
Vancouver & Adaptive DBN & 61.2\% & 0.2175 & 72.1\% & 10.0\% & 62.7\% & 0.0983 & 78.1\% & 58.2\%\\
& Adaptive DBN + TB & 80.5\% & 0.0421 & 87.8\% & 76.3\% & 82.7\% & 0.0484 & 88.1\% & 73.2\%\\
& Adaptive DBN + TB + TS & 88.9\% & 0.0322 & 93.0\% & 83.1\% & {\bf 95.4\%} & 0.0056 & {\bf 96.0\%} & {\bf 94.6\%}\\ \hline
Montreal & Adaptive DBN & 64.6\% & 0.2759 & 81.5\% & 6.0\% & 62.3\% & 0.0976 & 69.6\% & 55.6\%\\
& Adaptive DBN + TB & 83.1\% & 0.0639 & 92.9\% & 75.3\% & 83.4\% & 0.0057 & 86.3\% & 79.5\%\\
& Adaptive DBN + TB + TS & 91.4\% & 0.0356 & 95.9\% & 86.2\% & {\bf 97.3\%} & 0.0048 & {\bf 98.1\%} & {\bf 96.5\%}\\ 
\hline
\end{tabular}
\end{center}
\end{table*}

\begin{figure*}[tbp]
  \centering
  \subfigure[London, Adaptive DBN + TB + TS + Enclave]{\includegraphics[scale=1.20]{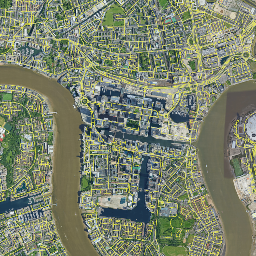}\label{fig:london_ts_tb_enclave}}
  \subfigure[Louisville, Adaptive DBN + TB + TS + Enclave]{\includegraphics[scale=1.20]{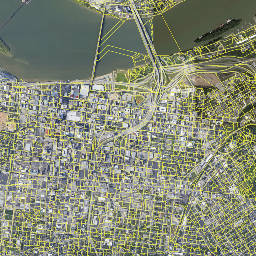}\label{fig:louisville_ts_tb_enclave}}
  \caption{Detection results with enclave search}
  \label{fig:result_roadtracer_ts_enclave}
\end{figure*}

Furthermore, some experimental results as shown in Table \ref{tab:result_roadtracer_ts_enclave2} and Fig.~\ref{fig:result_roadtracer_ts_enclave2}-Fig.~\ref{fig:result_roadtracer_ts_enclave3} for cities not mentioned in the paper are included in the Appendix.
}

\section{Detection of available roads in landslide disaster}
\label{sec:disaster}
\subsection{Experiment Environment}
In this section, we challenged to apply our proposed model to the detection of available roads in landslide by natural disaster in this section, because rapid detection of the available roads that are not broken by landslide is an important task to obtain a way of transportation and rescue residents even when a natural disaster is occurred. In the experiments, the satellite images in Hiroshima, Japan in July 2018, were used, where a heavy rainfall disaster was actually occurred. The two sets of images before and after the disaster were obtained from Geospatial Information Authority of Japan \cite{kokudo}. The image was divided into six areas from A to F and the road map detection was conducted for each image by using the proposed TS model. The size of each area is about 750 square mm in $1,024 \times 1,024$ pixels. Fig.~\ref{fig:map} shows the rough map around the areas A-F. The distance between each area is about 2 to 12km. The areas A-E are surrounded by the sea or mountains, where huge damage by the landslide was actually occurred in the disaster. F is the urban area and huge damage was not occurred in the disaster. There are few candidates for moving paths between each area. Fig.~\ref{fig:system_overview} shows an example of system implementation using our proposed model. As shown in Fig.~\ref{fig:system_overview}, a satellite or an aerial photo image that an aircraft collected is processed by an embedded device and the detected roads are visualized.

In this experiment, firstly, we tested whether our model can detect available roads in two sets of satellite images before and after the disaster, or not. Secondly, the inference speed of our model was verified on a small embedded edge device such as Jetson series computers. It is assumed that the model will be used to detect available roads from satellite images or aerial photographs in a certain observation point in real time. Although the choice using a GPU cloud computing is available, our model is expected to work on the small embedded device without GPU, because DBN is typically smaller network structure than CNN. From the website \cite{PASCO}, one aerial image can be obtained every 1.35 seconds. Therefore, we tested whether the inference speed of our model was faster 1.35 seconds on the embedded device for real-time inference, or not.

\begin{figure*}[tbp]
  \centering
  \subfigure[Area A (before the disaster)]{\includegraphics[scale=0.3]{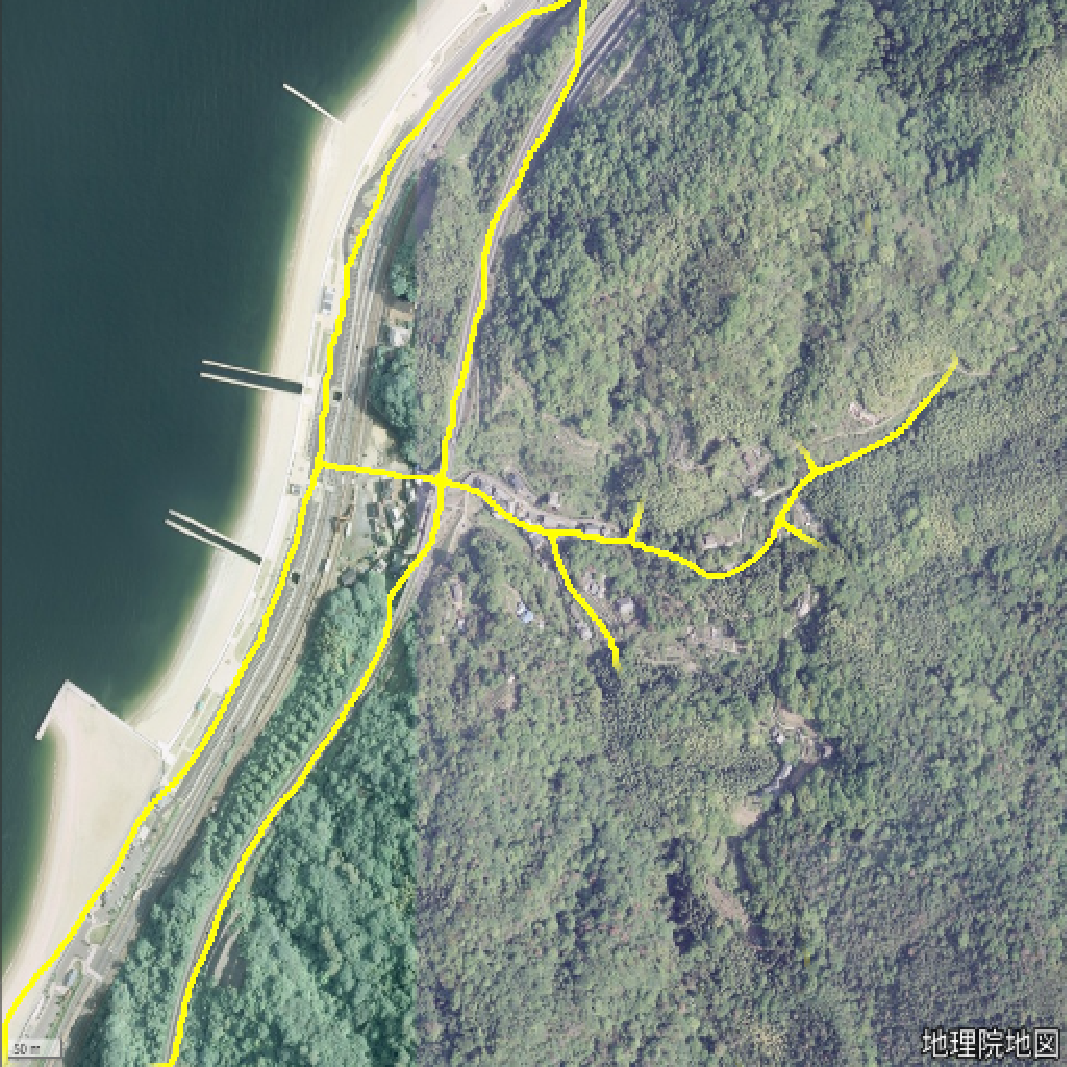}\label{fig:saka_before_result}}
  \subfigure[Area B (before the disaster)]{\includegraphics[scale=0.3]{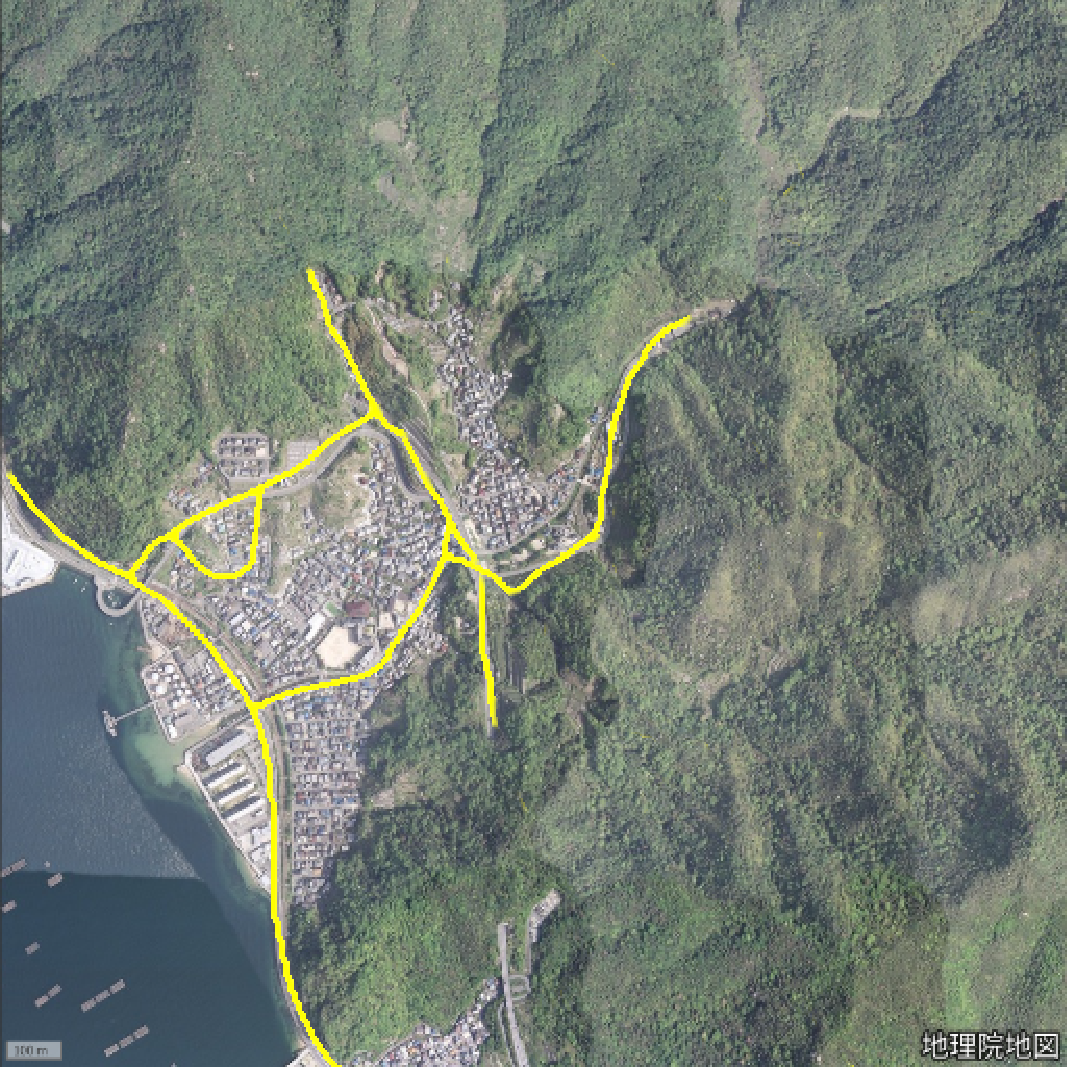}\label{fig:koyaura_before_result}}
  \subfigure[Area C (before the disaster)]{\includegraphics[scale=0.3]{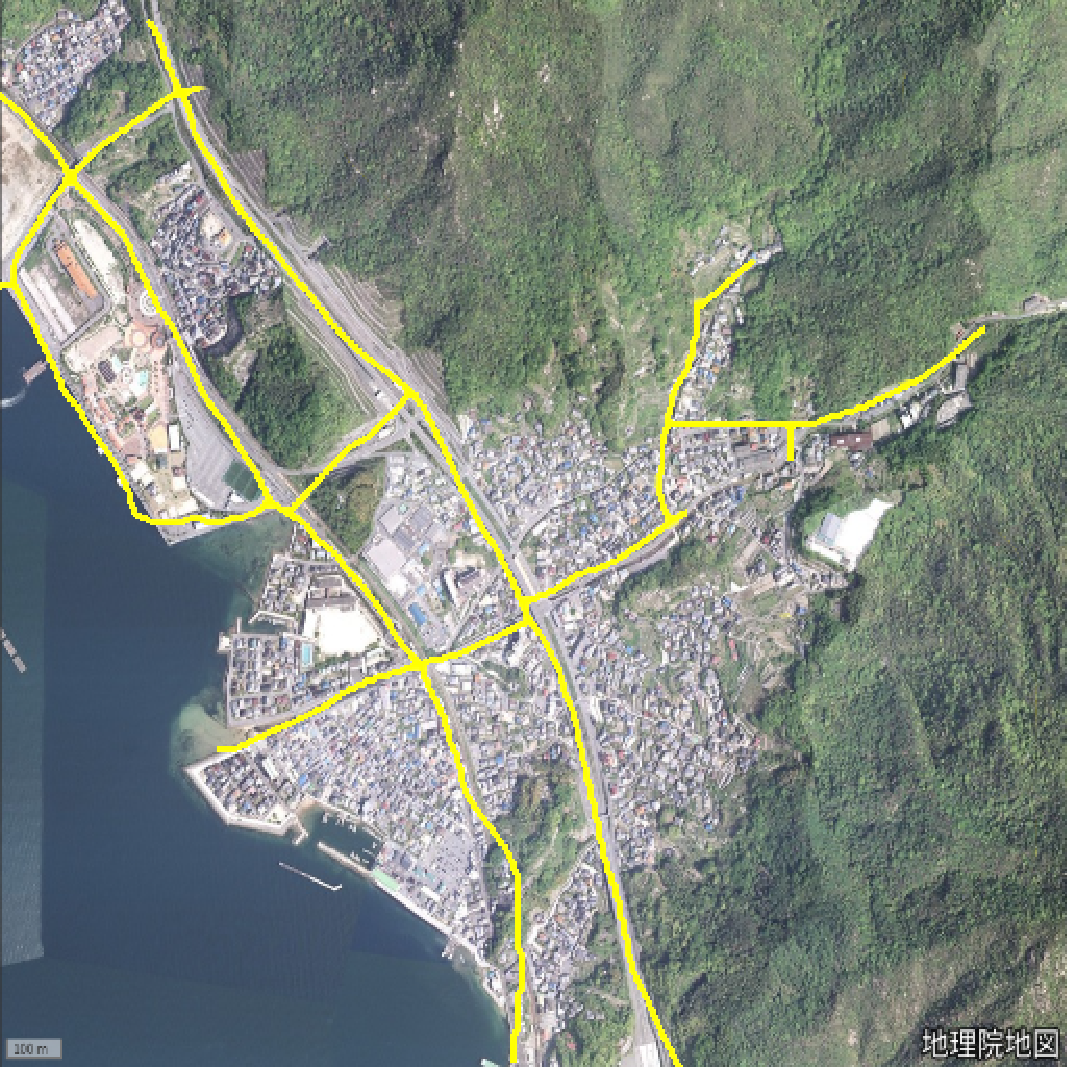}\label{fig:tenno_before_result}}
  \subfigure[Area A (after the disaster)]{\includegraphics[scale=0.3]{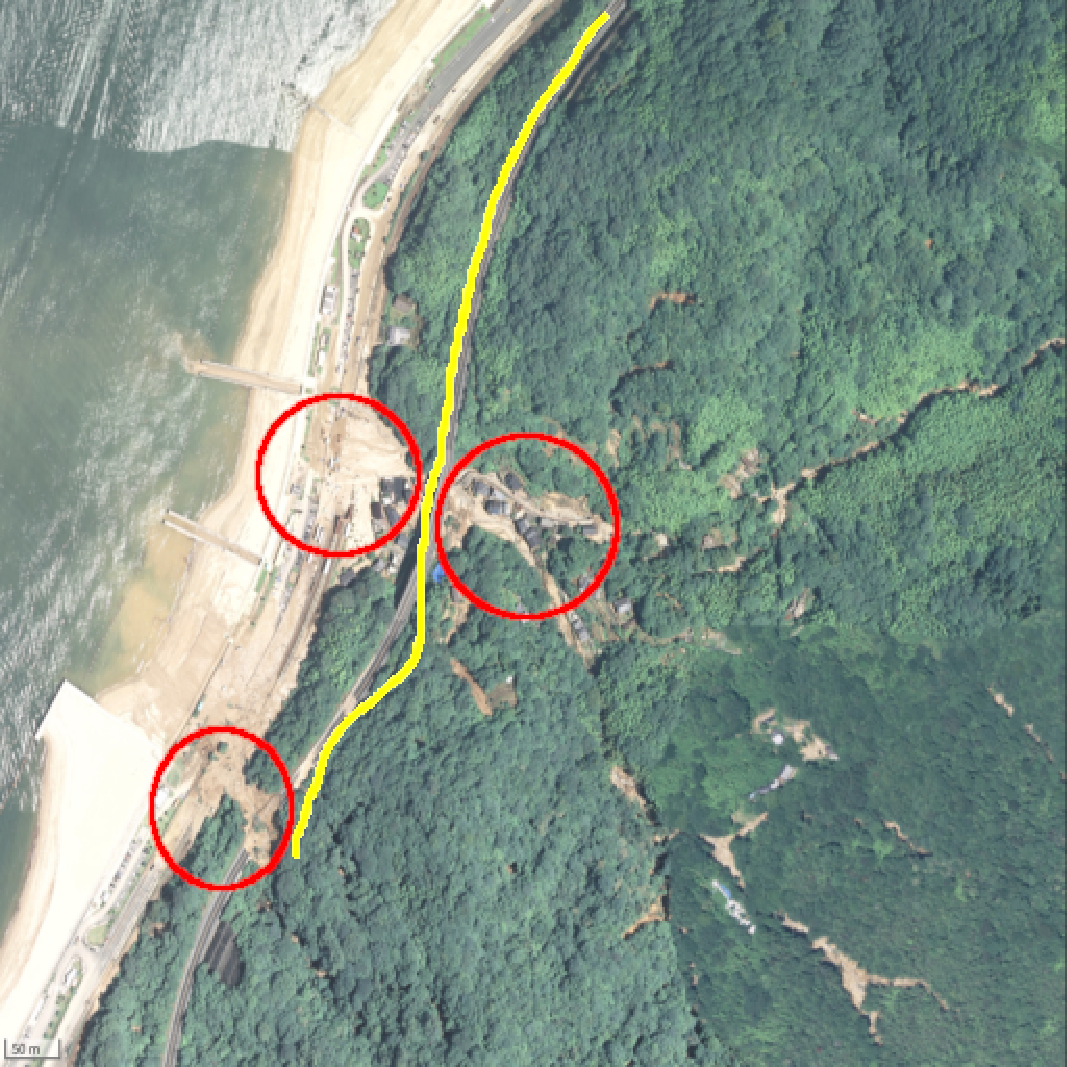}\label{fig:saka_after_result}}
  \subfigure[Area B (after the disaster)]{\includegraphics[scale=0.3]{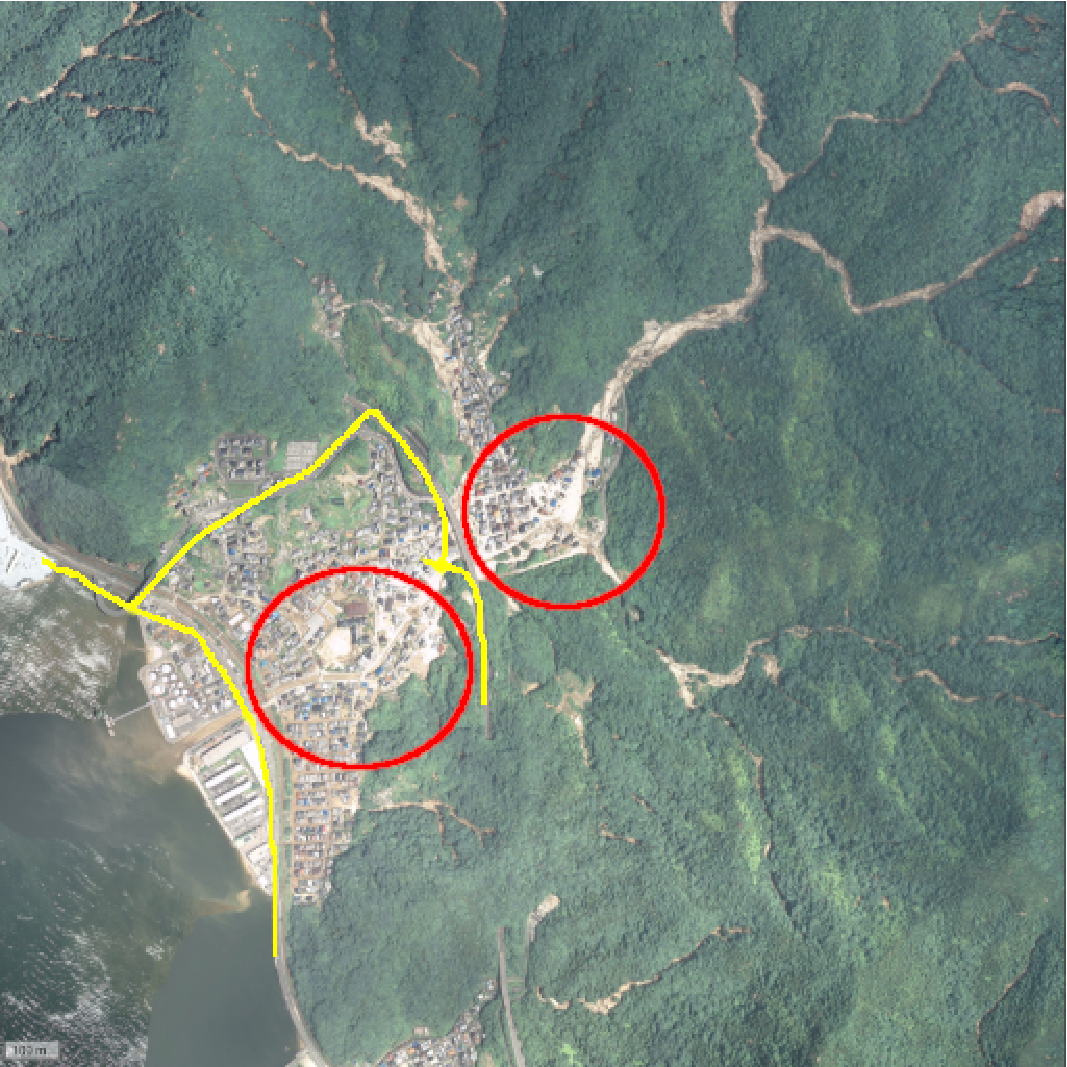}\label{fig:koyaura_after_result}}
  \subfigure[Area C (after the disaster)]{\includegraphics[scale=0.3]{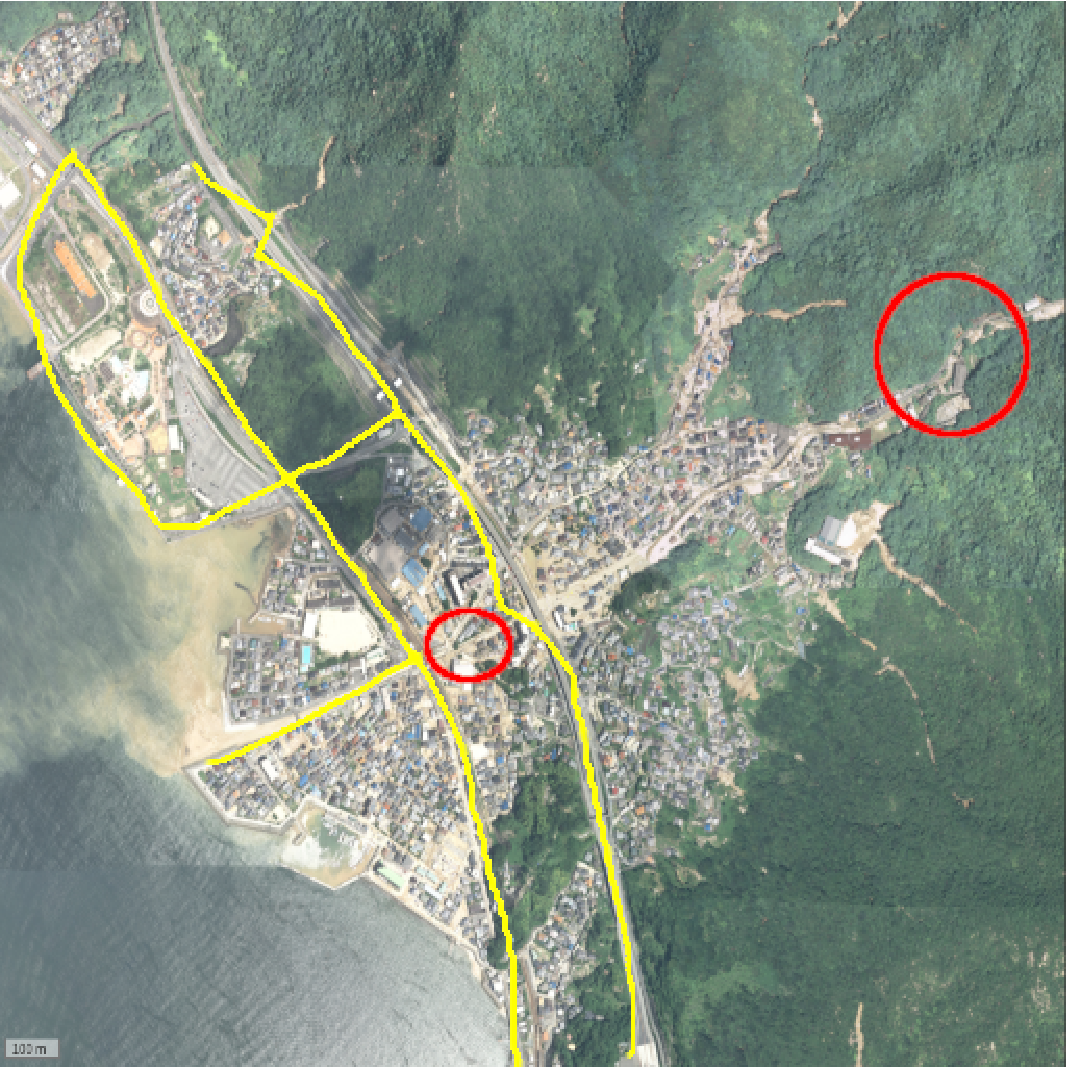}\label{fig:tenno_after_result}}
  \subfigure[Area D (before the disaster)]{\includegraphics[scale=0.3]{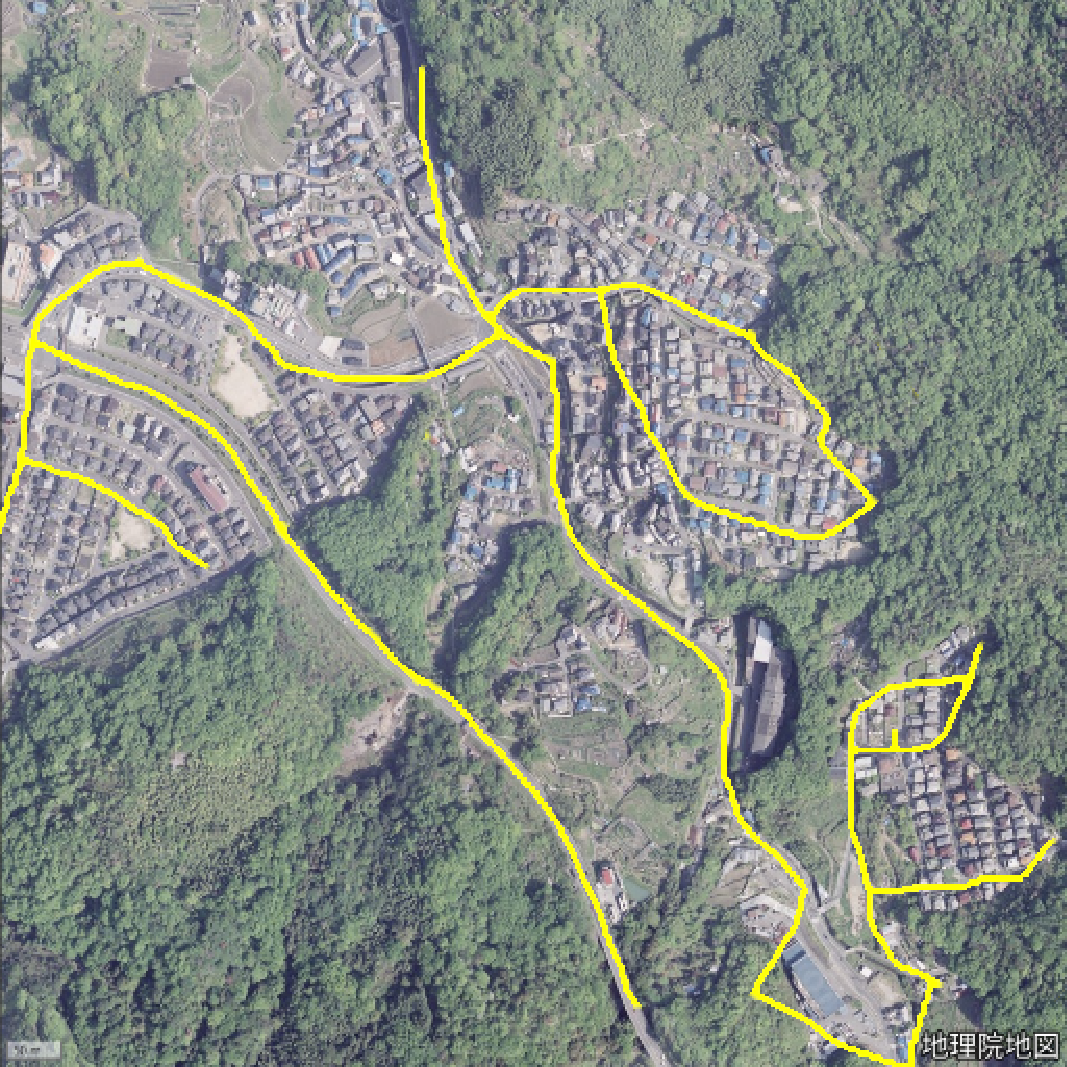}\label{fig:yano_before_result}}
  \subfigure[Area E (before the disaster)]{\includegraphics[scale=0.3]{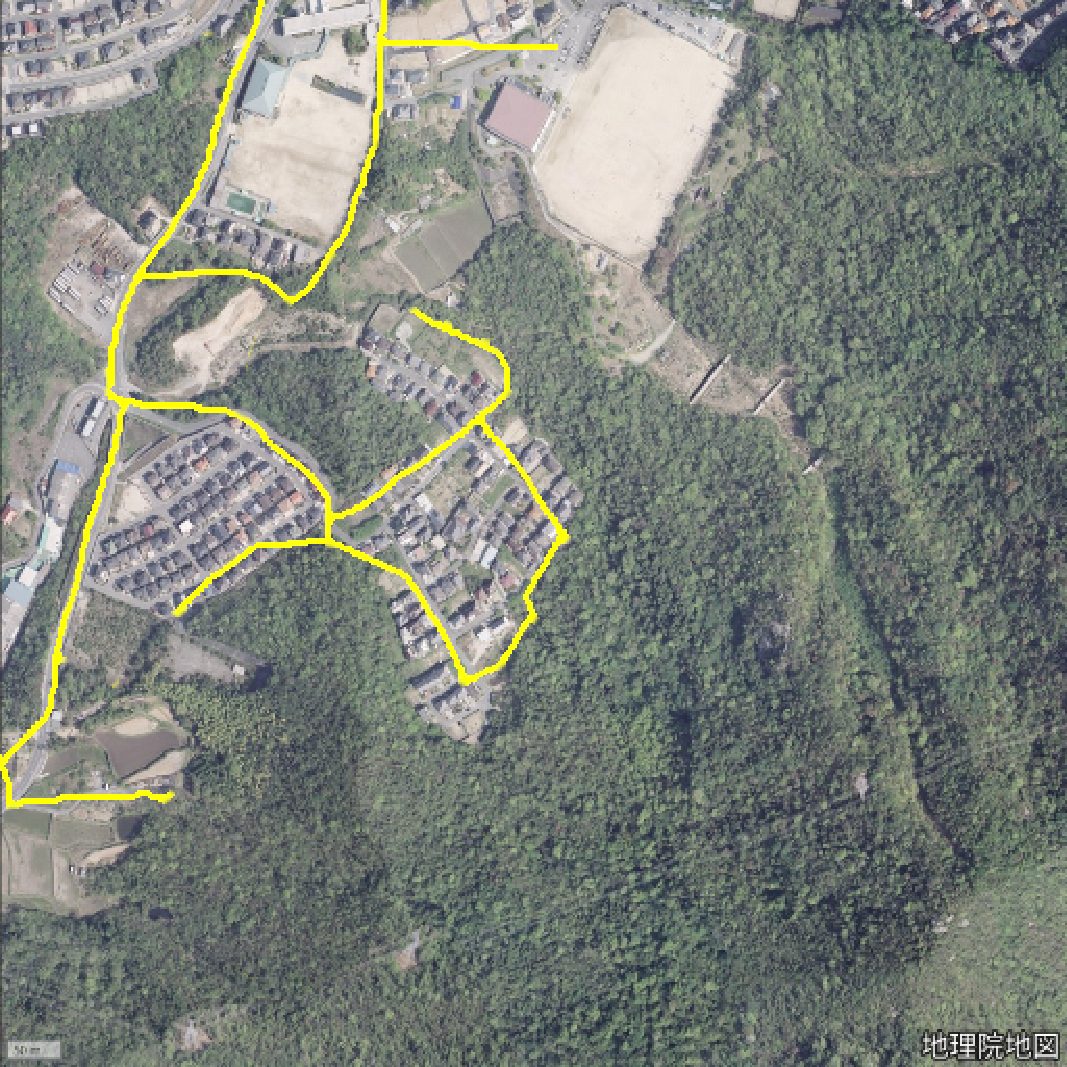}\label{fig:kumano_before_result}}
  \subfigure[Area F (before the disaster)]{\includegraphics[scale=0.3]{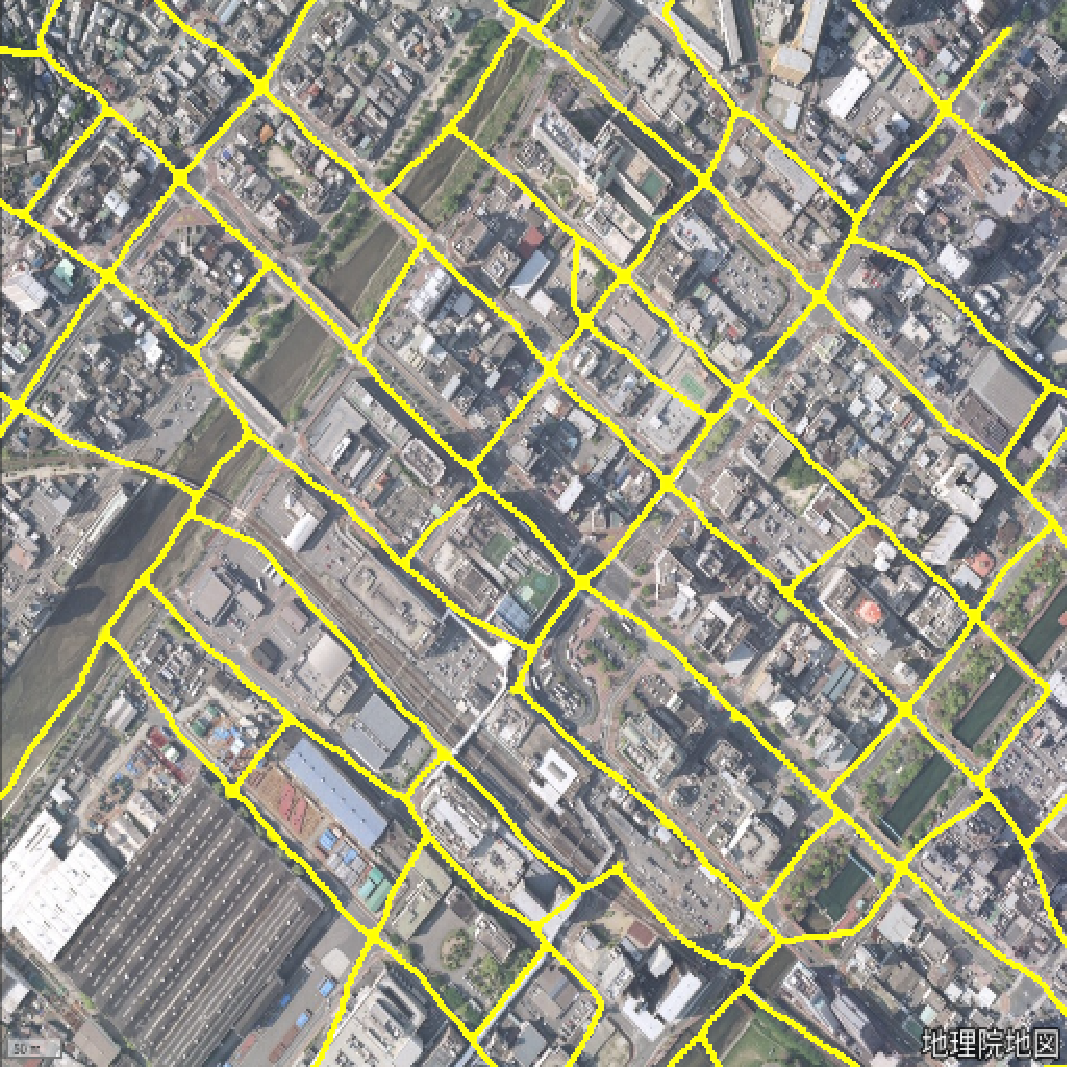}\label{fig:kure_before_result}}  
  \subfigure[Area D (after the disaster)]{\includegraphics[scale=0.3]{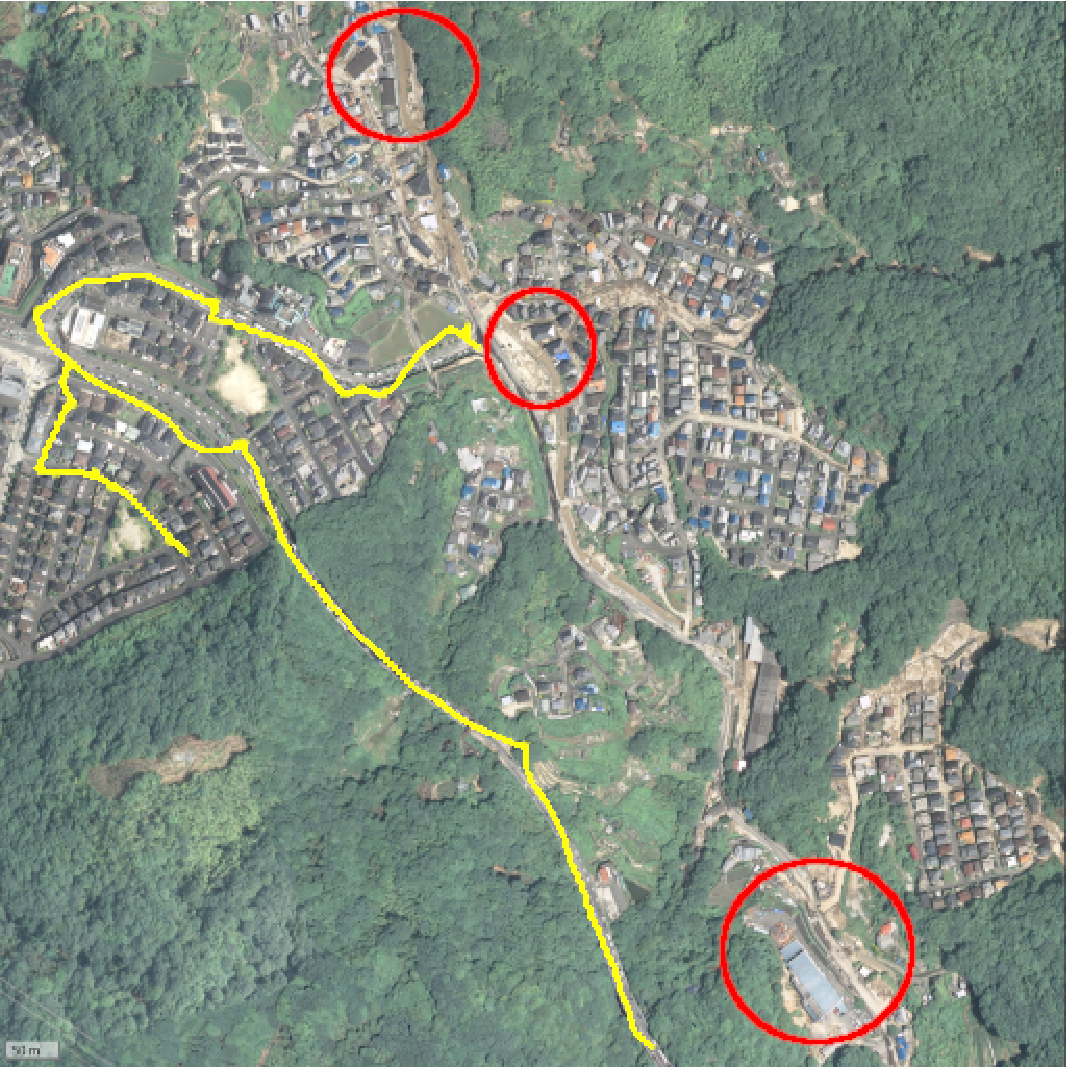}\label{fig:yano_after_result}}
  \subfigure[Area E (after the disaster)]{\includegraphics[scale=0.3]{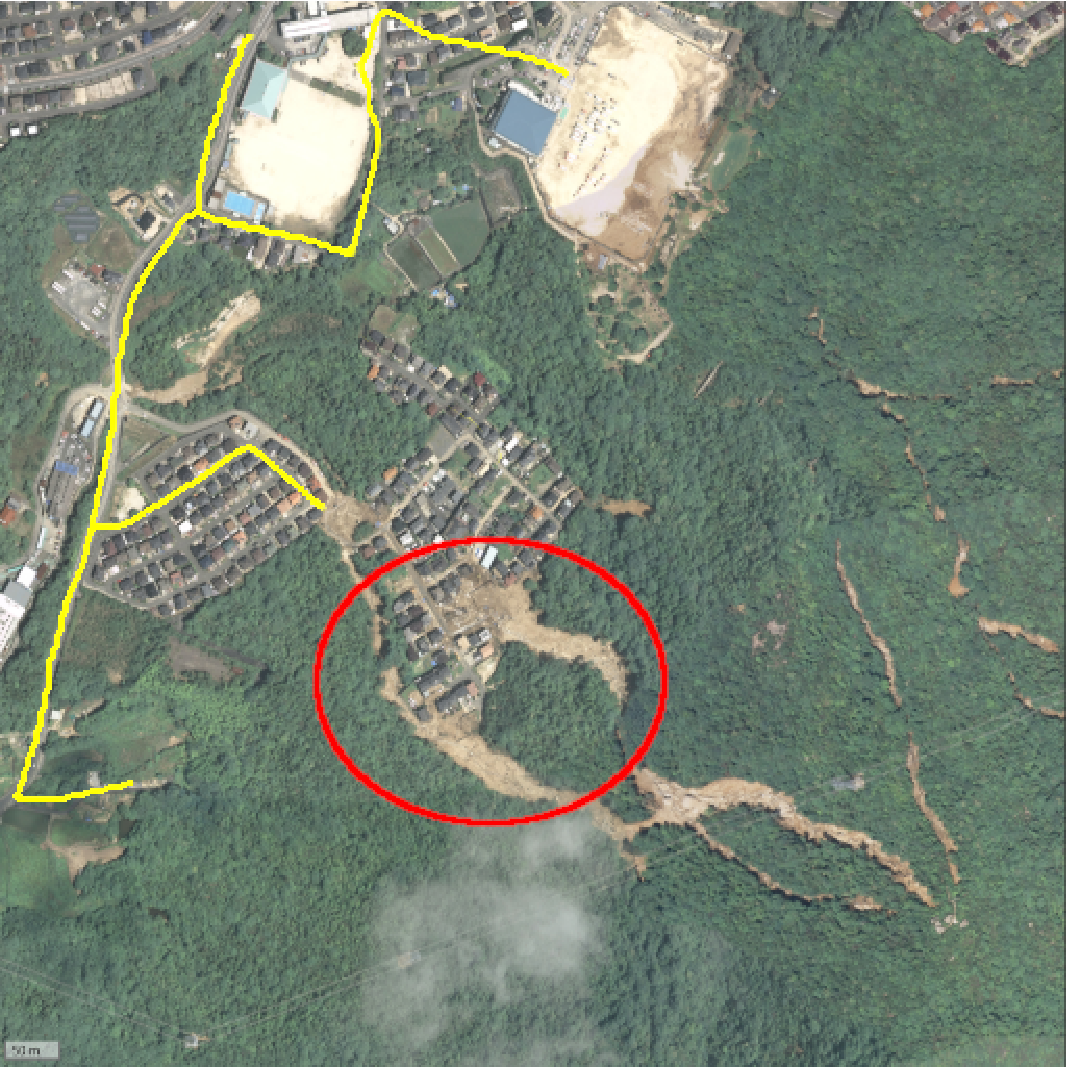}\label{fig:kumano_after_result}}
  \subfigure[Area F (after the disaster)]{\includegraphics[scale=0.3]{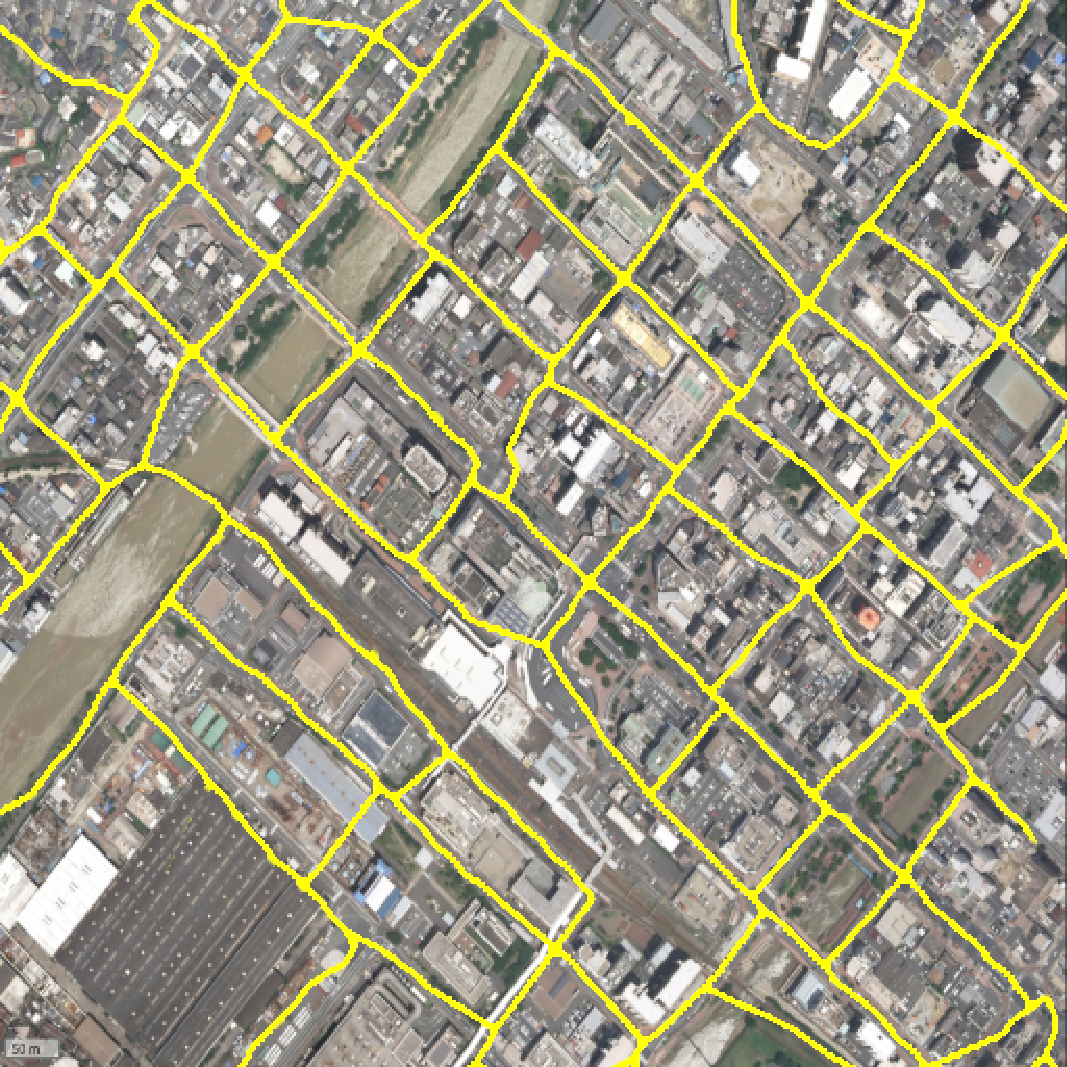}\label{fig:kure_after_result}}  
  \vspace{-3mm}
  \caption{Detection result before and after the disaster}
  \label{fig:result_disaster}
\end{figure*}

\subsection{Experiment Results}
Firstly, we tested whether our model can detect available roads in two sets of satellite images before and after the disaster, or not. Fig.~\ref{fig:result_disaster} shows the detection results before and after the disaster for the six areas. The red circles indicate areas that was not detected after the disaster due to the landslide. Table~\ref{tab:result_disaster_diff} shows the difference of the number of detected vertices before and after the disaster for six areas. We can see that the number of detected vertexes after the disaster is lower than that before the disaster because the connectivity between the roads could be broken by the disaster. From the results, some roads that were detected before the disaster were no longer detected after the disaster, except the area F because the terrible landslide was not occurred in the area.

\begin{figure}[tbp]
\centering
\includegraphics[scale=0.6]{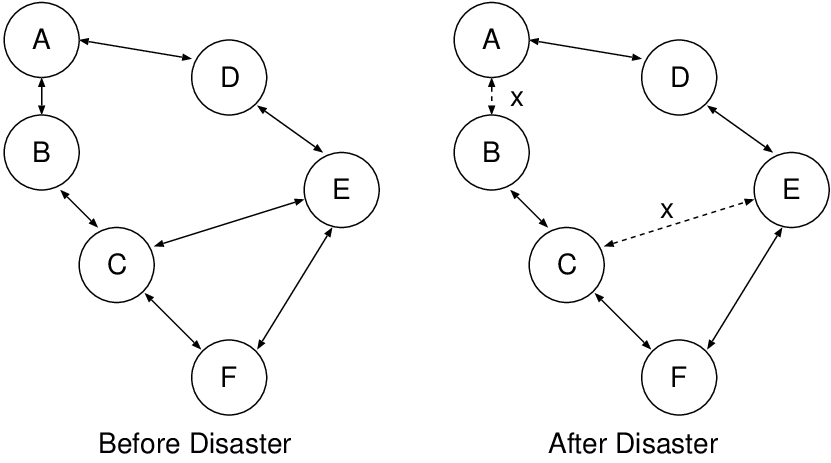}
\vspace{-5mm}
\caption{Available roads}
\label{fig:available_roads}
\vspace{-5mm}
\end{figure}

Therefore, the available roads after the disaster were extracted by calculating the difference of the two results. Fig.~\ref{fig:available_roads} shows the geographical relations among the six areas. The arrow means the available connection of roads between two areas. Before the disaster, there were seven connections in the six areas as shown in Fig.~\ref{fig:available_roads}. However, the detection results showed that the two of seven connections were not available after the disaster. The disconnection from A to B corresponds the not detected roads at lower left in Fig.\ref{fig:saka_after_result}. The disconnection from C to E corresponds the not detected roads at center to right direction (mountain) in Fig.\ref{fig:tenno_after_result}. The disconnection of the two roads showed that the bypass between the corresponding two areas was not able to use and then the evacuation or the supplies were suspended. At the time the disaster was actually occurred, the information of such available roads were manually checked by residents and shared by SNS or Google Map. That information and the detected available roads in this experiments matched. In conclusion, our model was able to detect the available roads in the satellite images after the disaster.

\begin{table}[tbp]
\caption{Available vertices after the disaster}
\label{tab:result_disaster_diff}
\centering
\begin{tabular}{c|c}
\hline 
Area & Detected vertices (after / before the disaster) \\ \hline
A & 13 / 42  \\ 
B & 28 / 39 \\
C & 28 / 40 \\ 
D & 21 / 42  \\  
E & 25 / 35  \\ 
F & 80 / 82  \\ 
\hline
\end{tabular}
\end{table}

Secondly, the inference speed of our model was verified on Jetson Xavier NX. For comparison, the original RoadTracer which consists of 17 convolutional layers was used. In addition, a small size of our model was constructed as the lightweight deep learning model to realize fast inference on the embedded edge device. In previous researches \cite{Kamada20_TENCON, Kamada22_Springer}, the fine-tuning method was developed, which can remove unnecessary neurons from the trained DBN model as pruning. After the training, frequency of activating neurons at each layer for given input is calculated. If unnecessary or inactivated neurons not contributing to output are found, the corresponding neurons are removed.

Table~\ref{tab:network_sotructure} shows the acquired network structure with/without the fine-tuning method. ``No. of neurons'' means the number of hidden neurons from input to output layers in six hidden layers. From Table~\ref{tab:network_sotructure}, the trained model had six hidden layers with and w/o fine-tuning. Generally, lower layers (close to input) tend to represent abstract feature of input patterns, while higher layers (close to output) tend to represent concrete features by stacking the hierarchical structure in deep learning. By applying the fine-tuning, about 100 neurons were removed from higher layers, not lower layers in the trained model. Although these neurons were constructed by the process of stacking the lower layers, they became unnecessary neurons for this image dataset.

Table~\ref{tab:inference_speed} shows the inference speed for each disaster area with and without GPU on Jetson Xavier NX in the three models. ``FPS'' means frames per second, where the size of a frame is $512 \times 512$ pixels. ``Searching Time(s)'' means seconds until search ends for an image of each disaster area with $1,024 \times 1,024$ pixels as shown in Fig.~\ref{fig:available_roads}. From Table~\ref{tab:inference_speed}, the searching time of the Adaptive DBN was about twice as fast as the original RoadTracer when using GPU, and seven times faster when using CPU. By fine-tuning the Adaptive DBN, the inference speed was slightly faster than before the fine-tuning (about 13\% improvement when using GPU, 16\% improvement when using CPU), without reducing the detection accuracy. It should be noted that the inference speed when using CPU is about only three times slower than when using GPU for the Adaptive DBN, while about 10 times slower for the original RoadTracer. Assuming one aerial image can be obtained every 1.35 seconds and use of CPU, the model that can realize real-time inference (faster than 1.35 seconds) was only the Adaptive DBN with and without the fine-tuning. Although, real-time inference can be realized on even the original RoadTracer if large resource with GPU is available, the proposed small deep learning model will be effective use for such urgent situations. Therefore, our proposed model showed higher detection accuracy and faster inference speed on CPU than the original RoadTracer.

\begin{table}[tbp]
\caption{Acquired network structure by fine-tuning}
\vspace{-5mm}
\label{tab:network_sotructure}
\begin{center}
\begin{tabular}{l|l}
\hline
         & No. of neurons  \\ \hline
Adaptive DBN w/o fine-tuning  & $710-521-410-213-135-40$  \\ 
Adaptive DBN with fine-tuning & $710-521-378-150-101-32$  \\ 
\hline
\end{tabular}
\end{center}
\end{table}

\begin{table*}[tbp]
\caption{Inference speed for each disaster area ($1,024 \times 1,024$ pixels)}
\vspace{-5mm}
\label{tab:inference_speed}
\begin{center}
\begin{tabular}{l|r|r|r|r}
\hline
&  \multicolumn{2}{c|}{CPU} & \multicolumn{2}{c}{GPU} \\ \cline{2-5} 
\multicolumn{1}{c|}{Model} & Searching Time(s) & FPS & Searching Time(s) & FPS \\ \hline
Original RoadTracer           & 6.365 & 0.63 & 0.624 & 6.41 \\
Adaptive DBN w/o fine-tuning  & 1.029 & 3.89 & 0.322 & 12.44 \\
Adaptive DBN with fine-tuning & 0.863 & 4.64 & 0.278 & 14.37 \\
\hline
\end{tabular}
\end{center}
\end{table*}

\section{Discussion}
\label{sec:discussion}
In remote sensing using deep learning, segmentation and LULC classification are famous approaches to find multiple objects from global region in satellite images \cite{Hong21A,Hong21B}. While these methods are segmentation approach, RoadTracer is another approach to extract only roads using graph search algorithm according to connectivity between nodes in local region of the satellite or aerial images. Recent success models to extend RoadTracer focused the training method of connectivity of roads \cite{Bahl22,Zhou18,Mei21,Zhang23}.

Although LULC classification and the road extraction methods are not comparable on quantitative manner because of different problems, LULC classification has an advantage to recognize multiple kinds of objects including roads. Whereas, the road extraction method has an advantage not to necessary require high-resolution satellite images with multi channels, this is, an aerial photography with RGB is available for training and inference. In addition, the road extraction method is more effective than LULC classification for the detection of available roads in disaster that our model tackled. In such a case that many buildings, roads, bridges, and so on, are broken by disaster, the search method that can determine connectivity between roads is more superior to segmentation. However, the original RoadTracer has less efficiency of search then often stops early due to local loop in no roads area. Therefore, the taboo search was implemented to enhance the exploration ability and the image recognition power was improved by TS model in this paper. Moreover, our proposed model realized faster inference speed than the original RoadTracer. The real-time inference on the GPU embedded device helps us to rapidly obtain a way of transportation and rescue residents when a disaster is actually occurred. Therefore, our proposed model is expected to be the most suitable for detecting available roads in disaster situations, considering its accuracy and computational efficiency.

Finally, the pros, cons, and limitation of our model are mentioned. Our proposed Adaptive DBN has the self-adaptation algorithm of the network structure for given training data, then it showed higher classification accuracy than the existing CNN models on several image benchmark datasets, as reported in \cite{Kamada18_Springer}. Whereas, CNN has an advantage over the DBN in terms of the model visualization, as well as Explainable AI (XAI), since CNN has convolutional filters to extract feature maps of the image. Moreover, the proposed TS model improved the image recognition power of Adaptive DBN as described in the experiments, since it can represent ambiguous patterns or fluctuated labels due to multiple annotators by using two or more student models. However, as the limitation, the TS model could not classify the cases if unguessable information from given input is included in the annotator's judgment based on external knowledge. For such cases, a multi-modal deep learning model using not only images but also other kinds of data such as text, numeric value, and so on, should be required.

\section{Conclusion}
\label{sec:conclusion}
RoadTracer is a generation method of a road map on the ground surface from aerial photograph data. In this paper, we proposed a novel method of RoadTracer using the TS based Adaptive DBN. The TS model is an ensemble learning model with one parent and multiple child models. The cross validation test showed the TS model achieved higher classification accuracy than the previous methods for the image benchmark datasets CIFAR-10 and CIFAR-100. Moreover, the TS model was also applied to the RoadTracer and the detection accuracy was improved from 40.0\% to 89.0\% on average in the seven cities. In addition, we challenged to apply our method to the detection of available roads when landslide by natural disaster is occurred. In conclusion, our model was implemented on the small embedded edge device by the fine-tuning method for fast inference and it was able to detect the available roads in the satellite images after the disaster in Japan. Such automatic detection method is expected to rapidly obtain a way of transportation and rescue residents when a disaster is actually occurred, since the recent unexpected weather or disaster have caused huge damage to not only mountainous zone but also urban zone, and the traditional method obtains such information by residents in the manner of manual check and SNS sharing. In future, a more efficient graph search algorithm will be developed by adding unsearchable space to the taboo list according to the image recognition result. The proposed model will be evaluated on the other disaster cases for further improvement in practical use.

\section*{Acknowledgments}
This work was supported by JSPS KAKENHI Grant Number 19K12142, 19K24365, and obtained from the commissioned research by National Institute of Information and Communications Technology (NICT), JAPAN.

\begin{IEEEbiography}[{\includegraphics[width=1in,height=1.25in,clip,keepaspectratio]{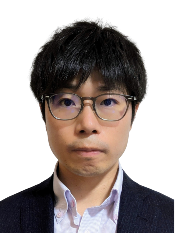}}]{Shin Kamada}
is an Associate Professor at the Graduate School of Information Sciences, Hiroshima City University. He received his ME and DE in Management and Information System from Prefectural University of Hiroshima in 2014, and Graduate School of Information Sciences, Hiroshima City University in 2019, respectively. He engaged in the theoretical and practical research in the research field of computational intelligence including Deep Learning.
\end{IEEEbiography}

\begin{IEEEbiography}[{\includegraphics[width=1in,height=1.25in,clip,keepaspectratio]{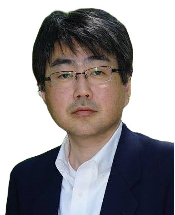}}]{Takumi Ichimura}
is a Professor at the Faculty of Regional Development, Prefectural University of Hiroshima, Japan. He received his ME and DE in Engineering at Toin University of Yokohama in 1994 and 1997, respectively. His current research interests include computational intelligence such as deep learning, and their applications in medical informatics and cognitive sciences.
  
\end{IEEEbiography}

\appendix
{\color{red}
\section{Experimental results with Enclave search}
Table \ref{tab:result_roadtracer_ts_enclave2} and Fig.~\ref{fig:result_roadtracer_ts_enclave2}-\ref{fig:result_roadtracer_ts_enclave3} show the experimental results obtained using the improved search system.

\begin{table*}[tbp]
\caption{Detection accuracy for other cities}
\vspace{-3mm}
\label{tab:result_roadtracer_ts_enclave2}
\begin{center}
\begin{tabular}{l|l|r|r|r|r}
\hline
 & & \multicolumn{4}{|c}{Enclave Search} \\ \cline{3-6}
City & Model & Ave. & Std. & Max & Min \\ \hline
Boston & Adaptive DBN & 78.6\% & 0.0071 & 79.4\% & 77.4\%\\
& Adaptive DBN + TB & 91.5\% & 0.0064 & 92.4\% & 90.4\%\\
& Adaptive DBN + TB + TS & {\bf 97.8\%} & 0.0052 & {\bf 98.4\%} & {\bf 97.2\%} \\\hline
Chicago & Adaptive DBN & 72.2\% & 0.0088 & 73.5\% & 71.2\%\\
& Adaptive DBN + TB & 84.8\% & 0.0078 & 86.3\% & 83.5\%\\
& Adaptive DBN + TB + TS & {\bf 94.8\%} & 0.0093 & {\bf 95.5\%} & {\bf 93.5\%}\\ \hline
Denver & Adaptive DBN & 75.4\% & 0.0077 & 76.5\% & 74.1\%  \\
& Adaptive DBN + TB & 90.4\% & 0.0062 & 91.0\% & 88.9\%  \\
& Adaptive DBN + TB + TS & {\bf 97.3\%} & 0.0036  & {\bf 97.9\%} & {\bf 96.5\%} \\ \hline
Kansas city & Adaptive DBN & 78.5\% & 0.0103 & 80.8\% & 66.8\%\\
& Adaptive DBN + TB & 87.2\% & 0.0094 & 88.5\% & 85.4\%\\
& Adaptive DBN + TB + TS & {\bf 92.2\%} & 0.0010 & {\bf 98.4\%} & {\bf 89.3\%}\\ \hline
Los Angles & Adaptive DBN & 77.3\% & 0.0090 & 78.2\% & 75.3\%\\
& Adaptive DBN + TB & 85.3\% & 0.0080 & 88.3\% & 79.3\%\\
& Adaptive DBN + TB + TS & {\bf 92.1\%} & 0.0046 & {\bf 92.5\%} & {\bf 91.2\%}\\ \hline
Paris & Adaptive DBN & 72.8\% & 0.0204 & 75.5\% & 69.6\%\\
& Adaptive DBN + TB & 83.2\% & 0.0123 & 86.3\% & 79.0\%\\
& Adaptive DBN + TB + TS & {\bf 97.6\%} & 0.0046 & {\bf 98.3\%} & {\bf 96.9\%}\\ \hline
Pittsburgh & Adaptive DBN & 76.1\% & 0.0718 & 71.5\% & 54.7\%\\
& Adaptive DBN + TB & 87.9\% & 0.0063 & 88.7\% & 87.1\%\\
& Adaptive DBN + TB + TS & {\bf 96.8\%} & 0.0038 & {\bf 97.4\%} & {\bf 96.3\%}\\ \hline
Saltlakecity & Adaptive DBN & 36.2\% & 0.0073 & 37.4\% & 35.2\%\\
& Adaptive DBN + TB & 66.6\% & 0.0085 & 68.6\% & 65.3\%\\
& Adaptive DBN + TB + TS & {\bf 90.1\%} & 0.0045 & {\bf 90.7\%} & {\bf 89.2\%}\\ \hline
San diego & Adaptive DBN & 85.5\% & 0.0892 & 86.2\% & 83.4\%\\
& Adaptive DBN + TB & 89.5\% & 0.0024 & 92.5\% & 84.2\%\\
& Adaptive DBN + TB + TS & {\bf 95.2\%} & 0.0064 & {\bf 95.7\%} & {\bf 94.3\%}\\ \hline
Toronto & Adaptive DBN & 88.5\% & 0.0261 & 89.0\% & 74.1\%\\
& Adaptive DBN + TB & 91.5\% & 0.0062 & 92.2\% & 88.9\%\\
& Adaptive DBN + TB + TS & {\bf 92.3\%} & 0.0073 & {\bf 93.0\%} & {\bf 91.1\%}\\ \hline \hline
Hiroshima-Koyaura & Adaptive DBN & 49.2\% & 0.0368 & 54.2\% & 42.4\%\\
& Adaptive DBN + TB & 78.1\% & 0.0350 & 82.2\% & 69.5\%\\
& Adaptive DBN + TB + TS & {\bf 96.9\%} & 0.0113 & {\bf 98.3\%} & {\bf 94.9\%}\\ \hline
Hiroshima-Saka & Adaptive DBN & 43.6\% & 0.1297 & 53.1\% & 29.6\%\\
& Adaptive DBN + TB & 67.0\% & 0.0236 & 76.4\% & 52.1\%\\
& Adaptive DBN + TB + TS & {\bf 95.0\%} & 0.0074 & {\bf 96.4\%} & {\bf 93.9\%}\\ \hline
Hiroshima-Tenno & Adaptive DBN & 44.7\% & 0.0130 & 51.6\% & 30.6\%\\
& Adaptive DBN + TB & 69.2\% & 0.0238 & 80.4\% & 31.8\%\\
& Adaptive DBN + TB + TS & {\bf 96.8\%} & 0.0011 & {\bf 98.2\%} & {\bf 94.5\%}\\ \hline

\hline
\end{tabular}
\end{center}
\end{table*}

\begin{figure*}[tbp]
  \centering
  \subfigure[New York, Adaptive DBN + TB + TS + Enclave]{\includegraphics[scale=1.00]{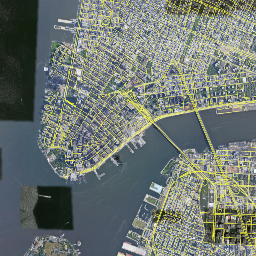}\label{fig:newyork_ts_tb_enclave}}
  \subfigure[Tokyo, Adaptive DBN + TB + TS + Enclave]{\includegraphics[scale=1.00]{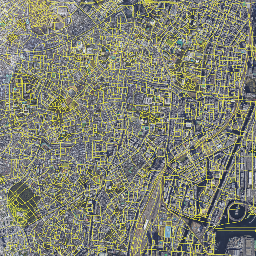}\label{fig:tokyo_ts_tb_enclave}}
  \subfigure[Amsterdam, Adaptive DBN + TB + TS + Enclave]{\includegraphics[scale=1.00]{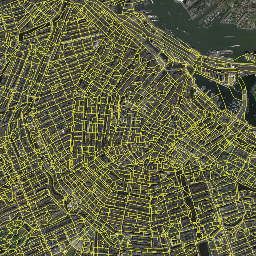}\label{fig:amsterdam_ts_tb_enclave}}
  \subfigure[Vancouver, Adaptive DBN + TB + TS + Enclave]{\includegraphics[scale=1.00]{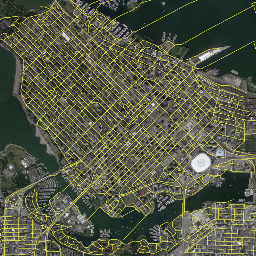}\label{fig:vancouver_ts_tb_enclave}}
  \subfigure[Montreal, Adaptive DBN + TB + TS + Enclave]{\includegraphics[scale=1.00]{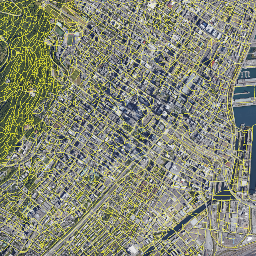}\label{fig:montreal_ts_tb_enclave}}
  \subfigure[Boston, Adaptive DBN + TB + TS + Enclave]{\includegraphics[scale=1.00]{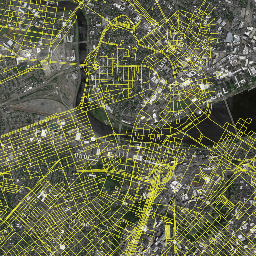}\label{fig:boston_ts_tb_enclave}}
%
  \subfigure[Chicago, Adaptive DBN + TB + TS + Enclave]{\includegraphics[scale=1.00]{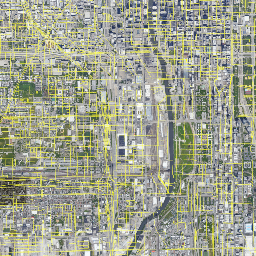}\label{fig:chicago_ts_tb_enclave}}
  \subfigure[Denver, Adaptive DBN + TB + TS + Enclave]{\includegraphics[scale=1.00]{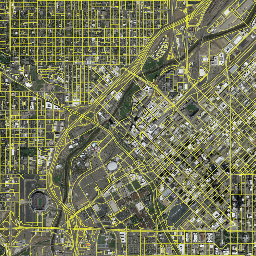}\label{fig:denver_ts_tb_enclave}}   
  \subfigure[Kansas City, Adaptive DBN + TB + TS + Enclave]{\includegraphics[scale=1.00]{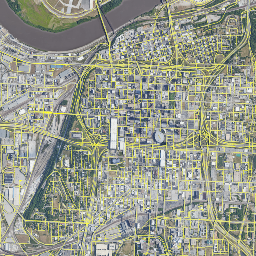}\label{fig:kansas_city_ts_tb_enclave}}
  \subfigure[Los Angesl, Adaptive DBN + TB + TS + Enclave]{\includegraphics[scale=1.00]{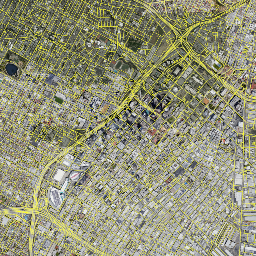}\label{fig:la_ts_tb_enclave}}
  \subfigure[Paris, Adaptive DBN + TB + TS + Enclave]{\includegraphics[scale=1.00]{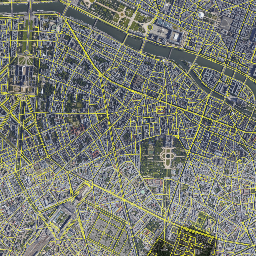}\label{fig:paris_ts_tb_enclave}}
  \subfigure[Pittsburgh, Adaptive DBN + TB + TS + Enclave]{\includegraphics[scale=1.00]{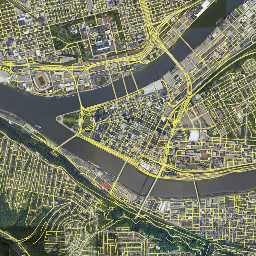}\label{fig:pittsburgh_ts_tb_enclave}} 
%
  \subfigure[Salt Lake City, Adaptive DBN + TB + TS + Enclave]{\includegraphics[scale=1.00]{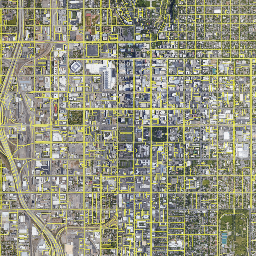}\label{fig:saltlakecity_ts_tb_enclave}}
  \subfigure[San Diego, Adaptive DBN + TB + TS + Enclave]{\includegraphics[scale=1.00]{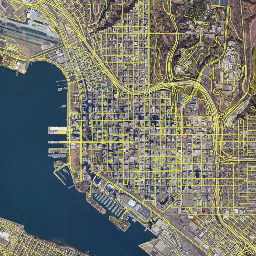}\label{fig:sandiego_ts_tb_enclave}}   
  \subfigure[Toronto, Adaptive DBN + TB + TS + Enclave]{\includegraphics[scale=1.00]{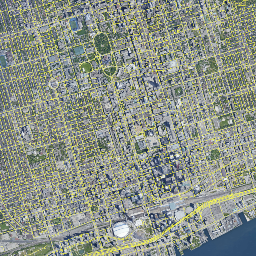}\label{fig:toronto_ts_tb_enclave}}
  \caption{Detection results with enclave search 2}
  \label{fig:result_roadtracer_ts_enclave2}
\end{figure*}

\begin{figure*}[!ht]
  \centering
  
  \subfigure[Hiroshima-Koyaura, Adaptive DBN + TB + TS + Enclave]{\includegraphics[scale=1.5]{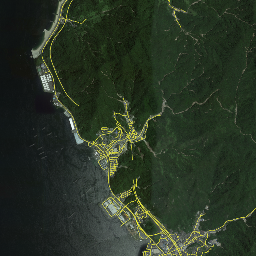}\label{fig:Koyaura_ts_tb_enclave}}
  \subfigure[Hiroshima-Saka, Adaptive DBN + TB + TS + Enclave]{\includegraphics[scale=1.5]{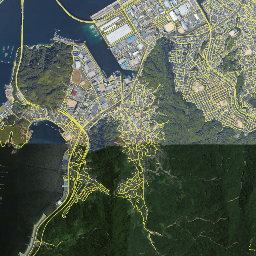}\label{fig:Saka_ts_tb_enclave}}
  \subfigure[Hiroshima-Tenno, Adaptive DBN + TB + TS + Enclave]{\includegraphics[scale=1.5]{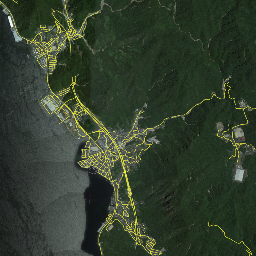}\label{fig:Tennno_ts_tb_enclave}}
  \caption{Detection results with enclave search 3}
  \label{fig:result_roadtracer_ts_enclave3}
\end{figure*}
}
\end{document}